\newcommand{\figpref}{Fig. }
\newcommand{\tabpref}{Table }

\newcommand{\secpref}{Section }
\documentclass[conference]{IEEEtran}
\usepackage{cite}

\ifCLASSINFOpdf
   \usepackage[pdftex]{graphicx}
   \graphicspath{{./}}
   \DeclareGraphicsExtensions{.pdf,.jpeg,.png}
\else
   \usepackage[dvips]{graphicx}
   \graphicspath{{./}}    
   \DeclareGraphicsExtensions{.eps}
\fi
\usepackage{multirow}
\usepackage{eqparbox}
\usepackage{mdwmath}
\usepackage{mdwtab}
\usepackage{amsmath}
\usepackage{amsfonts}
\usepackage{amssymb}
\usepackage{array}

\usepackage{threeparttable}

\ifCLASSOPTIONcompsoc
  \usepackage[caption=false,font=normalsize,labelfont=sf,textfont=sf]{subfig}
\else
  \usepackage[caption=false,font=footnotesize]{subfig}
\fi
\usepackage{fixltx2e}

\usepackage{stfloats}

\usepackage{tabularx}
\usepackage{color, soul}
\usepackage{longtable}
\usepackage{supertabular}
\usepackage{gensymb}
\usepackage{footnote}
\makesavenoteenv{tabular}
\makesavenoteenv{table}

\usepackage{acronym}
\usepackage{setspace}

\usepackage{fancyhdr}
\fancypagestyle{firstpage}{
	\fancyhf{}
	\fancyhead[L]{Under review of IPIN 2017}
}
\begin{document}

%
\title{Joint Positioning and Radio Map Generation \\Based on Stochastic Variational Bayesian Inference for FWIPS}

%


%
\author{\IEEEauthorblockN{Caifa Zhou\IEEEauthorrefmark{1},
Yang Gu\IEEEauthorrefmark{1}\IEEEauthorrefmark{2}}
\IEEEauthorblockA{\IEEEauthorrefmark{1}Institute of Geodesy and Photogrammetry, ETH Zurich\\
Stefano-Franscini-Platz 5, 8093, Zurich}
\IEEEauthorblockA{\IEEEauthorrefmark{2}College of Electronic Science and Engineering\\
National University of Defense Technology\\
Changsha, Hunan, P.R. China\\
Email: \{caifa.zhou, yang.gu\}@geod.baug.ethz.ch}}


\maketitle
\thispagestyle{firstpage}

\begin{abstract}
Fingerprinting based WLAN indoor positioning system (\acs{fwips}) provides a promising indoor positioning solution to meet the growing interests for indoor location-based services (e.g., indoor way finding or geo-fencing). \acs{fwips} is preferred because it requires no additional infrastructure for deploying an \acs{fwips} and achieving the position estimation by reusing the available \acs{wlan} and mobile devices, and capable of providing absolute position estimation. For \acf{fbp}, a model is created to provide reference values of observable features (e.g., signal strength from \acf{ap}) as a function of location during offline stage. One widely applied method to build a complete and an accurate reference database (i.e. \acf{rm}) for \acs{fwips} is carrying out a site survey throughout the \acf{roi}. Along the site survey, the readings of \acf{rss} from all visible \acsp{ap} at each \acf{rp} are collected. This site survey, however, is time-consuming and labor-intensive, especially in the case that the \acs{roi} is large (e.g., an airport or a big mall). This bottleneck hinders the wide commercial applications of \acs{fwips} (e.g., proximity promotions in a shopping center). To diminish the cost of site survey, we propose a probabilistic model, which combines \acf{fbp} and \acs{rm} generation based on \acf{svbi}. This \acs{svbi} based position and \acs{rss} estimation has three properties: i) being able to predict the distribution of the estimated position and \acs{rss}, ii) treating each observation of \acs{rss} at each \acs{rp} as an example to learn for \acs{fbp} and \acs{rm} generation instead of using the whole \acs{rm} as an example, and iii) requiring only one time training of the \acs{svbi} model for both localization and \acs{rss} estimation. These benefits make it outperforms the previous proposed approaches. We validate the proposed approach via experimental simulation and analysis. Compared to the \acs{fbp} approaches based on a \acf{snn}, \acf{dnn} and \acf{knn}, the proposed \acs{svbi} based position estimation outperforms them. The reduction of \acl{rmse} of the localization is up to 40\% comparing to that of \acs{snn} based \acs{fbp}. Moreover, the cumulative positioning accuracy, defined as the cumulative distribution function of the positioning errors, of the proposed \acs{fbp} and \acs{knn} are 92\% and 84\% within 4 $ \mathrm{m} $, respectively. The improvement of the positioning accuracy is up to 8\%. Regarding the performance of \acs{svbi} based \acs{rm} generation, it is comparable to that of the manually collected \acs{rm} and adequate for the applications, which require the room level positioning accuracy.  
\end{abstract}


%
\IEEEpeerreviewmaketitle

	\acrodef{roi}[RoI]{region of interest}
	\acrodef{fwips}[FWIPS]{fingerprinting based WLAN indoor positioning system}
	\acrodef{fbp}[FbP]{fingerprinting based positioning}
	\acrodef{ap}[AP]{access point}
	\acrodef{rss}[RSS]{received signal strength}
	\acrodef{wlan}[WLAN]{wireless local area network}
	\acrodef{rm}[RM]{radio map}
	\acrodef{wrt}[w.r.t.]{with respect to}
	\acrodef{rp}[RP]{reference point}
	\acrodef{tp}[TP]{test point}
	\acrodef{wifi}[WiFi]{wireless fidelity}
	\acrodef{knn}[$k\mathrm{NN} $]{$ k $ nearest neighbors}
	\acrodef{adagrad}[AdaGrad]{adaptive stochastic gradient descent}
	\acrodef{nn}[NN]{neural network}
	\acrodef{dnn}[DNN]{deep neural network}
	\acrodef{rmse}[RMSE]{root mean squared error}
	\acrodef{bm}[BM]{baseline model}
	\acrodef{relu}[ReLu]{rectified linear unit}
	\acrodef{mse}[MSE]{mean squared error}
	\acrodef{se}[SE]{squared error}
	\acrodef{tft}[TfT]{time for training}
	\acrodef{poe}[PoE]{patience of the early stopper}
	\acrodef{ae}[AE]{auto-encoder}
	\acrodef{snn}[SNN]{single layer neural network}
	\acrodef{bfgs}[BFGS]{Broyden-Fletcher-Goldfarb-Shannon}
	\acrodef{lm}[LM]{Levenberg Marquardt}
	\acrodef{ce}[CE]{cross Entropy}
	\acrodef{sdae}[SDAE]{stacked denoising auto-encoder}
	\acrodef{dlm}[DLM]{deep learning model}
	\acrodef{adam}[Adam]{adaptive moment estimation}
	\acrodef{adadelta}[AdaDelta]{adaptive Delta}
	\acrodef{rmsprop}[RMSprop]{RMS propagation}
	\acrodef{iid}[i.i.d.]{independent identical distributed}
	\acrodef{svbi}[SVBI]{stochastic variational Bayesian inference}
	\acrodef{rhs}[r.h.s]{right hand side}
	\acrodef{mcmc}[MCMC]{Monte Carlo Markov chain}
	\acrodef{mcs}[MCS]{Monte Carlo sampling}
	\acrodef{mle}[MLE]{maximum likelihood estimation}
	\acrodef{dlpm}[DLPM]{deep learning based positioning models}
	\acrodef{rfid}[RFID]{radio frequency identification}
	\acrodef{uwb}[UWB]{ultra wide band}
	\acrodef{ips}[IPS]{indoor positioning system}
	\acrodef{imu}[IMU]{inertial measurement unit}
	\acrodef{ts}[TS]{total station}
	\acrodef{atr}[ATR]{automatic target recognition}
	\acrodef{mle}[MLE]{maximum likelihood estimation}
	\acrodef{map}[MAP]{maximum a posteriori}
	\acrodef{pdr}[PDR]{pedestrian dead-reckoning}
	\acrodef{cpa}[CPA]{cumulative positioning accuracy}

\section{Introduction}\label{sec:introduction}
Fingerprinting based WLAN indoor positioning systems (\acsp{fwips}) have been attracting attention \cite{Adler2015} from both academia and industry in the last decades for their advantages in the following two aspects: they do not necessarily require special or additional infrastructure and they have limited error bounds because they {yield} absolute positions. While other \acfp{ips}, such as \acs{ips} based on \acf{rfid}, \acf{uwb}, supersonic (e.g., Crickets) or infrared signals (e.g., Active Badge), require deploying and operating dedicated infrastructure. \acsp{fwips} use existing \acs{wlan} \acfp{ap} and \acs{wifi} enabled devices (e.g., mobile phones, and tablets)\cite{1432143,hazas2006broadband,Youssef2008,liu2007survey}. Other infrastructure-free \acsp{ips} (e.g., \acl{pdr} based on the built-in \acfp{imu} of the mobile devices), they provide position changes while need to be {integrated} and the errors of positioning grow with time. \cite{1528431,4167810}.

Generally, an \acs{fwips} is realized using two stages: offline and online stage. During the offline stage a reference database representing the reference values of observable features as a function of location is created. This database is often named \acf{rm}). For creating this database, a site survey throughout the \acf{roi} usually has to be conducted. In this paper we assume that the \acs{rm} consists of a set of \acfp{rp} and the corresponding readings of \acfp{rss} from all visible \acsp{ap}. Because of changes of the indoor environment and configuration  of the \acs{wlan}, the offline mapping process needs to be carried out frequently to keep the \acs{rm} up-to-date. This time-consuming and labor-intensive site survey significantly impairs the widespread application of \acs{fwips} apart from academic research.

\begin{figure*}[!htb]
		\centering			
		\subfloat[Original \acs{rm}]{
			\label{subfig:orgrm}
			\includegraphics[height=4.5cm]{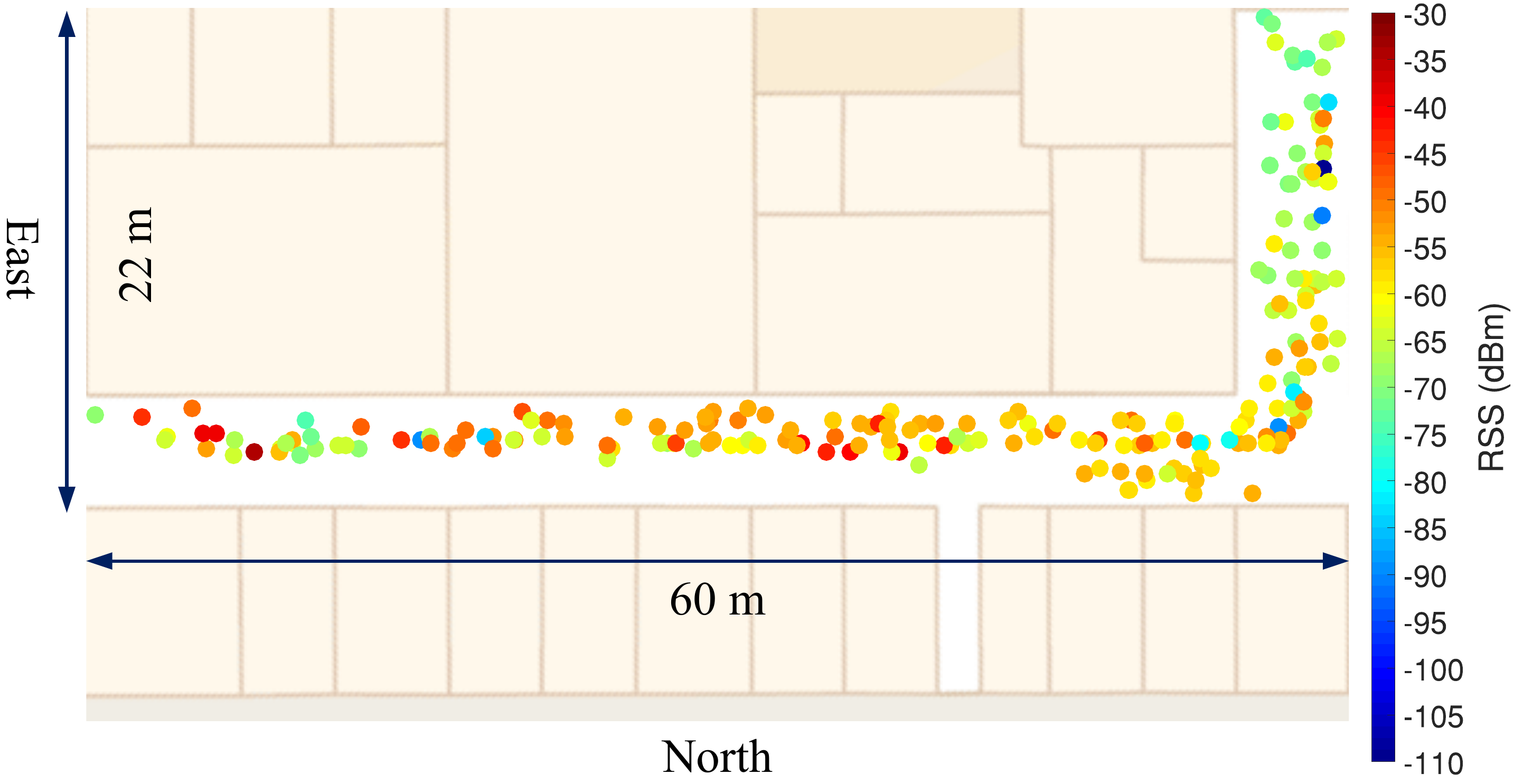}}
		\subfloat[Standardized original \acs{rm}]{
			\label{subfig:stdorgrm}
			\includegraphics[height=4.5cm]{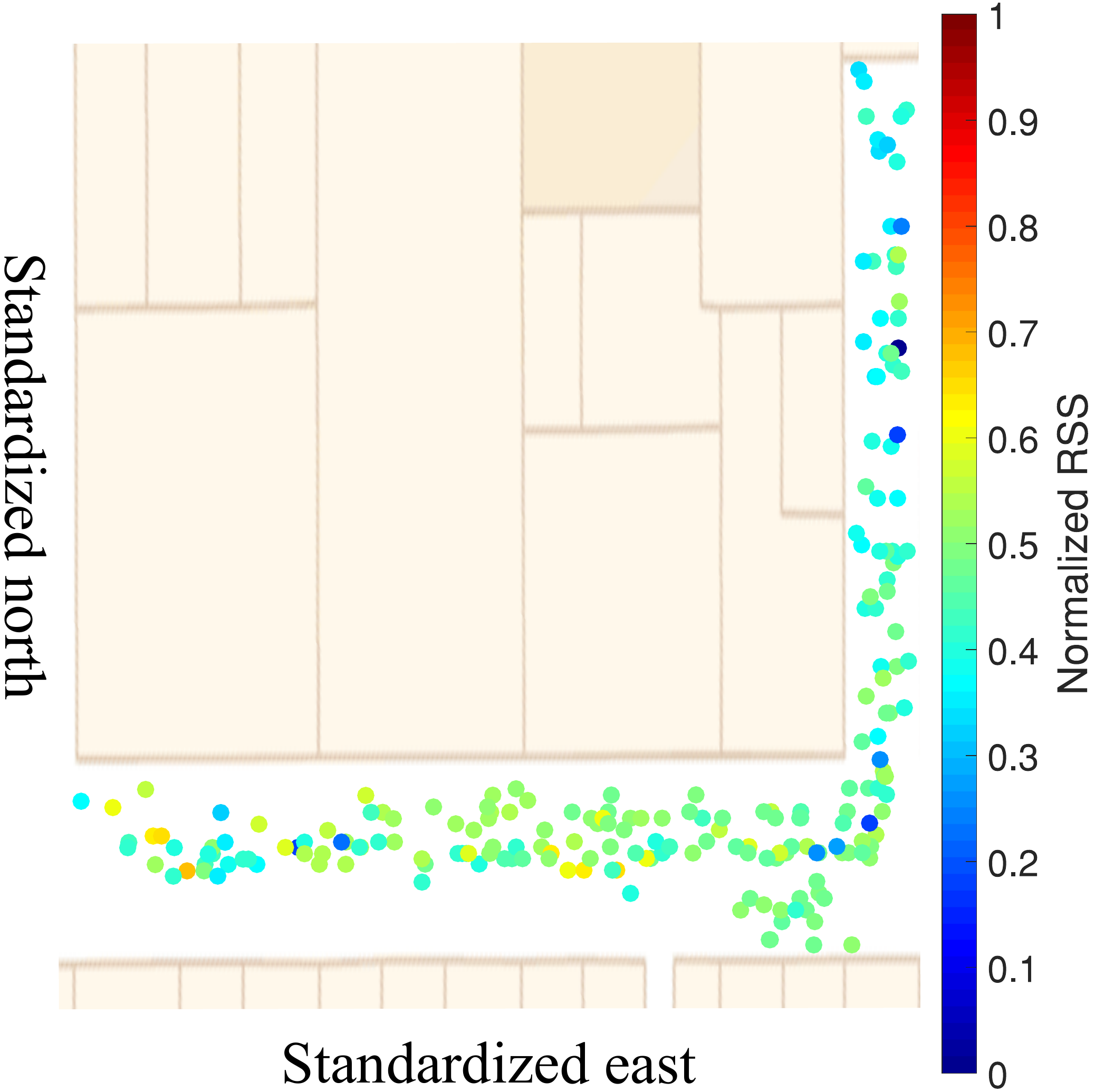}}\\
		\subfloat[Value of latent variable ($3^{\mathrm{rd}}$ dim.)]{
			\label{subfig:lat3}
			\includegraphics[height=4.5cm]{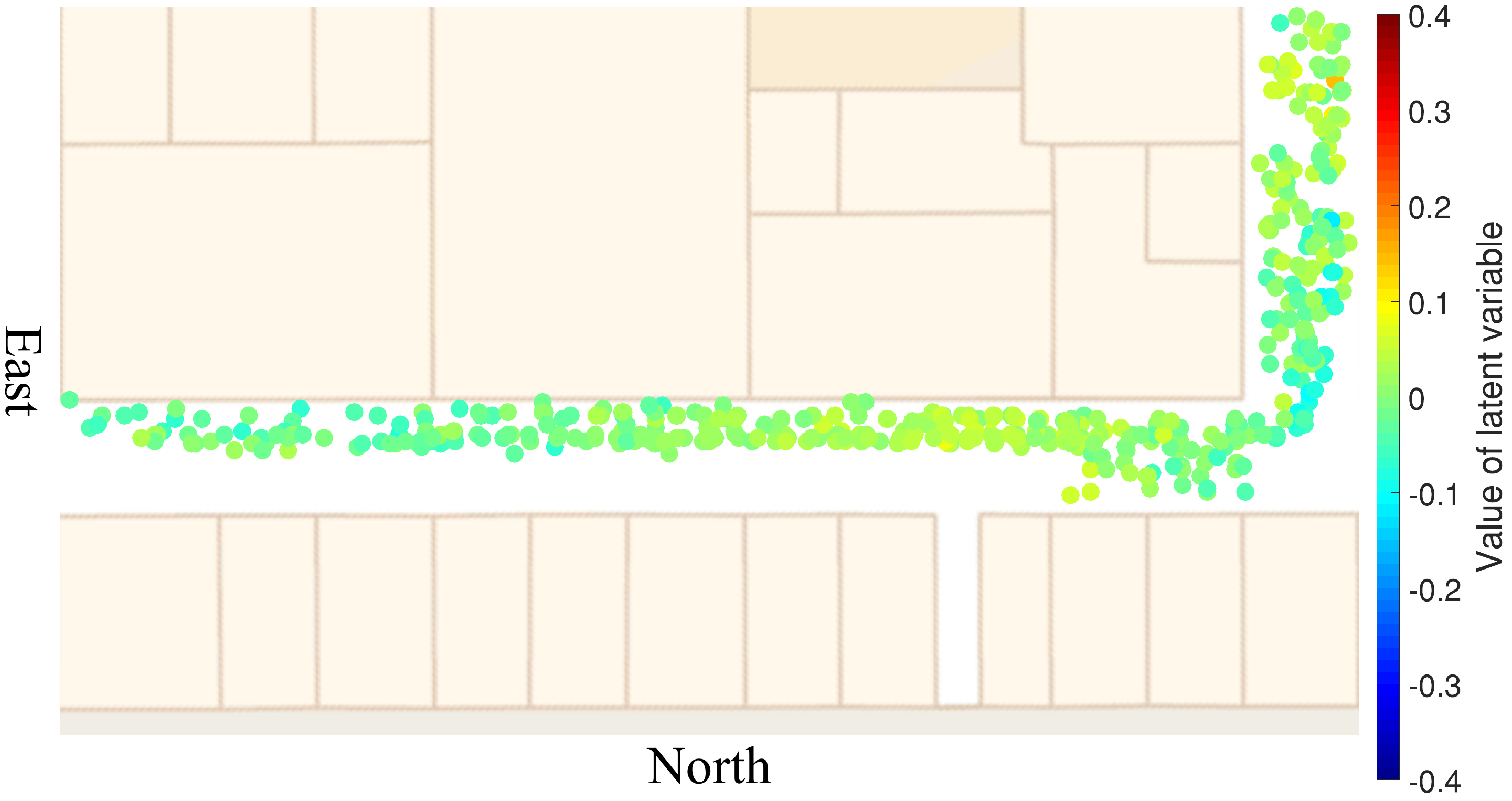}}
		\subfloat[Value of latent variable ($4^{\mathrm{th}}$ dim.)]{
			\label{subfig:lat4}
			\includegraphics[height=4.5cm]{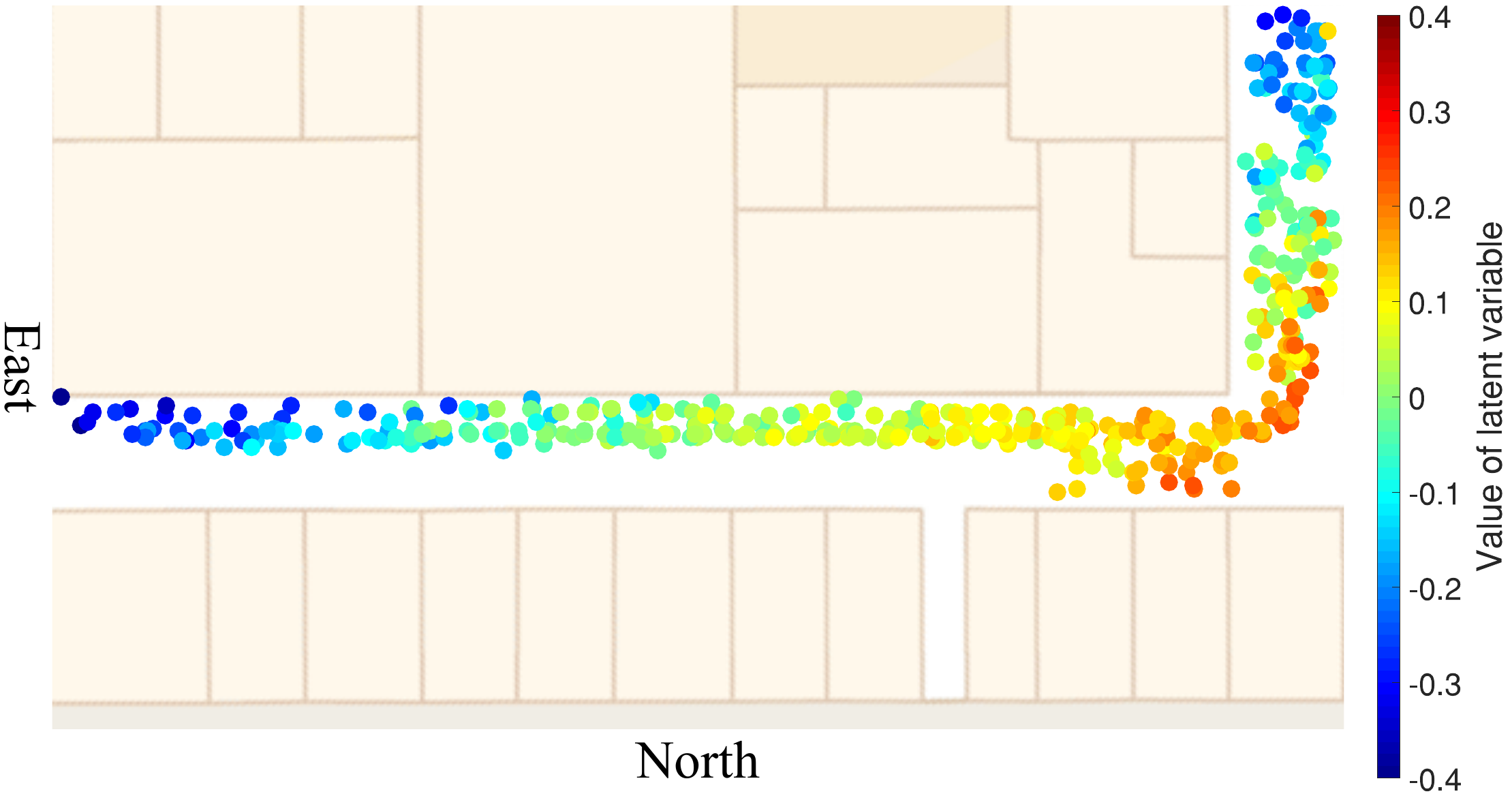}}\\
		\subfloat[Generated \acs{rm}]{
			\label{subfig:genrm}
			\includegraphics[height=4.5cm]{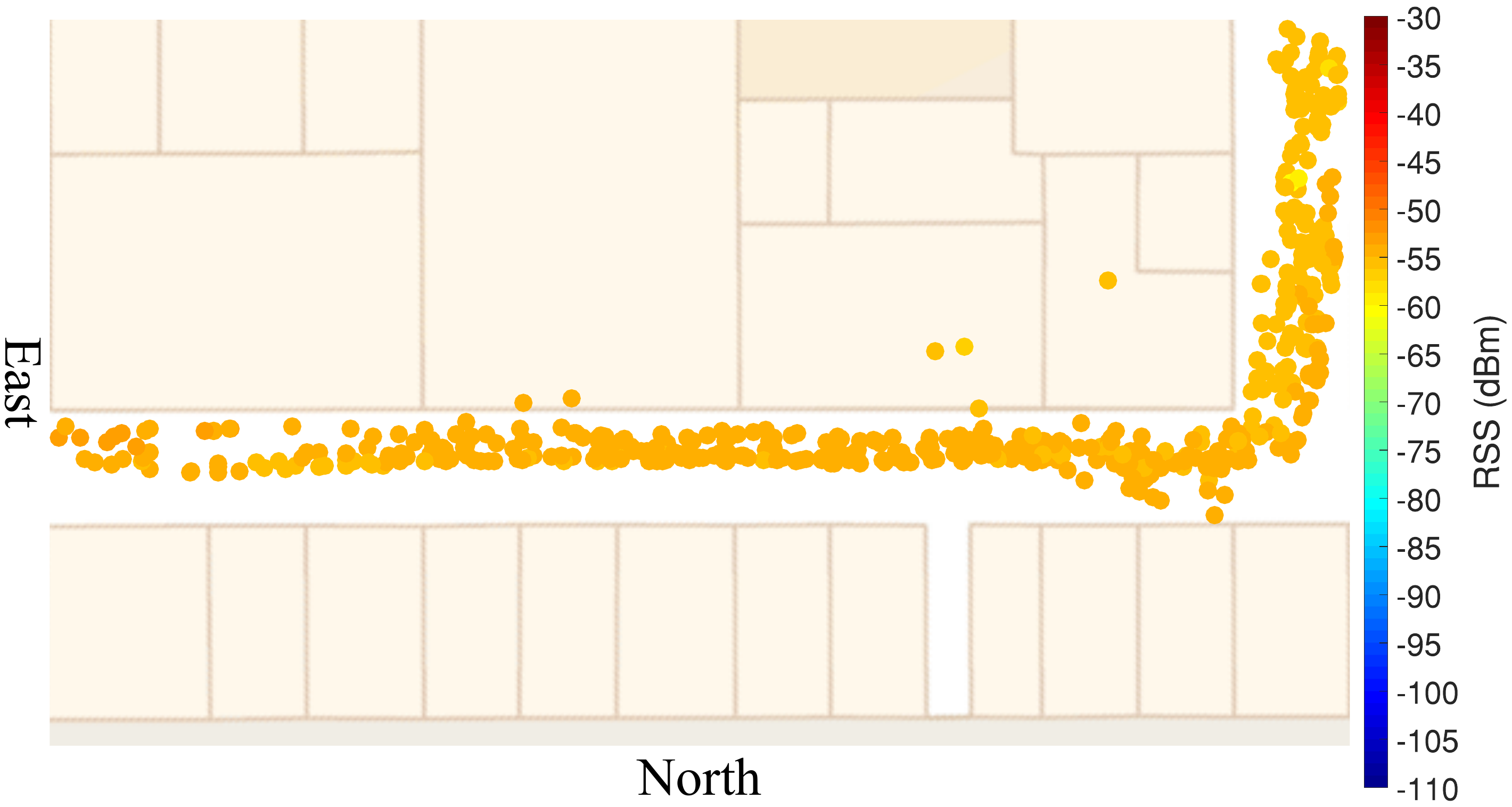}}
		\subfloat[Schematic of \acs{knn} ($k=3$) positioning using the generated \acs{rm}]{
			\label{subfig:knngraph}
			\includegraphics[height=4.5cm]{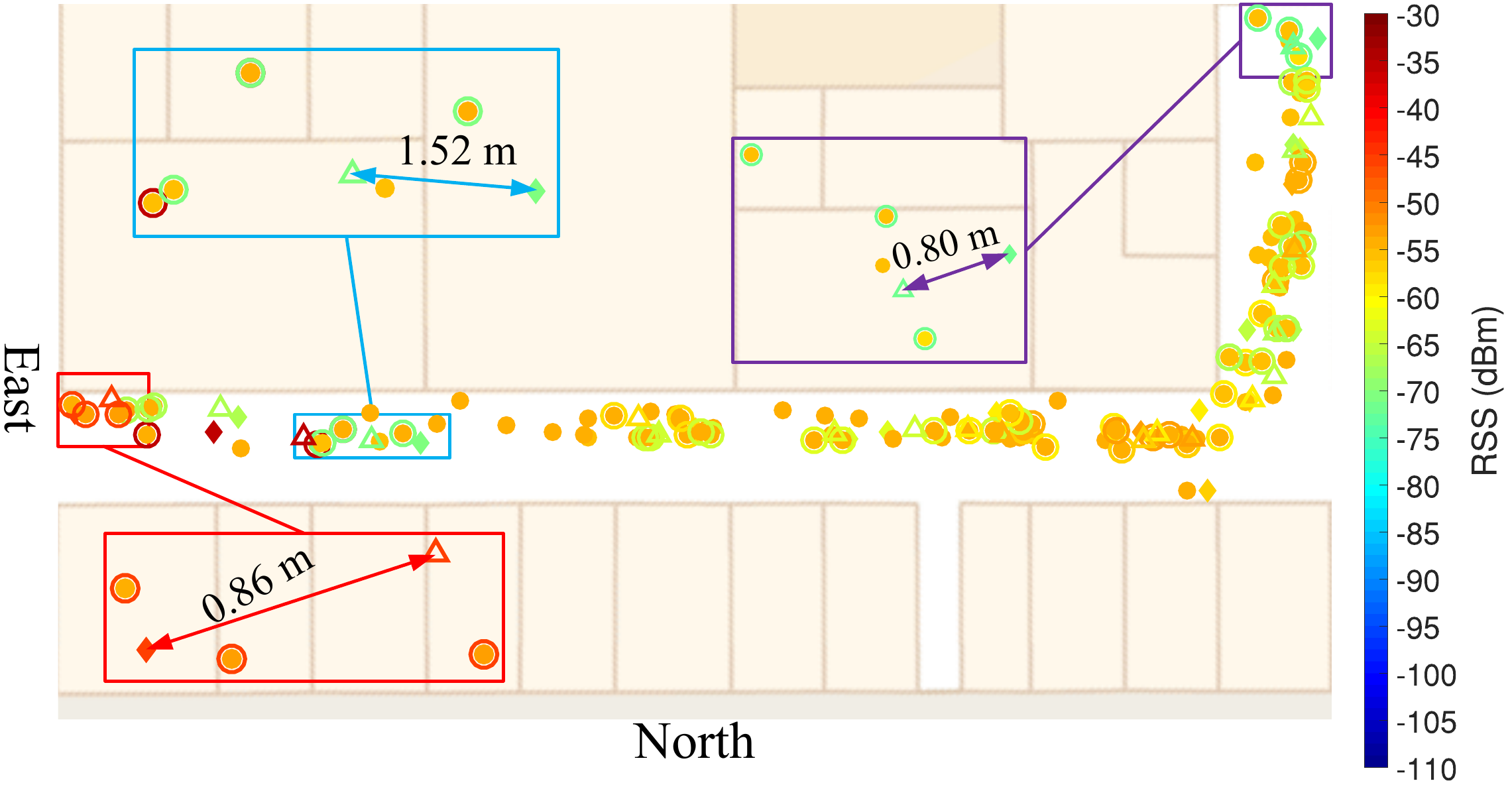}}
		\caption{An example of \acf{svbi} based joint \acs{fbp} and \acf{rm} generation. Except for \figpref\ref{subfig:stdorgrm}, the range of the \acs{roi} in other five figures are the same as \figpref\ref{subfig:orgrm}. The filled circle denotes the \acf{rp} and the filled color indicates the value of the respective variable depicted in each figure. In \figpref\ref{subfig:knngraph}, the meaning of the symbols are: i) the filled diamond depicts the ground truth of the \acf{tp}; ii) the found three nearest neighbors of each \acs{tp} are circled by a unfilled circle; iii) the unfilled triangle represents the estimated position of the \acs{tp} using \acs{knn} ($k=3$); and iv) the three enlarged parts show the positioning errors. Each of the enlarged part has the different enlarging scale.}
		\label{fig:examplesvbi}
	\end{figure*}

To reduce the time and labor of \acs{rm} collection and update, a mathematical model for both position estimation and \acs{rm} generation based on \acf{svbi} is proposed herein. \acs{svbi} is employed to project the \acsp{rss} from the high-dimensional \acs{rss}-space to a {much} lower-dimensional space of so-called latent variables (\figpref\ref{subfig:lat3} and \figpref\ref{subfig:lat4}), from which the position and corresponding \acs{rss} can then be simultaneously estimated. Via training \acs{svbi} based joint position and \acs{rss} estimation model using a collected \acs{rm}, it can achieve to \acs{rm} generation (\figpref\ref{subfig:genrm}) by implementing \acs{svbi} based position and \acs{rss} estimation model using a \acf{nn} which we train using Keras \footnote{Keras is provided via {https://github.com/fchollet/keras}. There is no paper about Keras. The advantages of Keras are i) built-in parallel programming on CPU and GPU; ii) stochastic gradient descent based backpropagation algorithms included; and iii) callbacks (e.g., early stopper and dropout) available for training the deep learning model and avoid the easy-over-fitting problem.}, a machine learning library for deep learning.

In \secpref\ref{sec:relatedWorks} we give an overview on selected previous work related to \acsp{fwips}, especially on publications focusing on \acf{fbp} using \acsp{nn}, and on joint \acs{fbp} and \acs{rm} generation. Fundamentals and formulations of \acs{fbp} and \acsp{nn} are presented in \secpref\ref{sec:fundamentals}. The proposed \acs{svbi} model is analyzed in \secpref\ref{sec:svbi}. An experimental analysis and a discussion of the proposed approach is given in \secpref\ref{sec:expanadis}.

\section{Related Work}\label{sec:relatedWorks}
We give a brief review on previously published \acs{fbp} approaches, especially the ones using \acsp{nn}. Then methods for \acs{rm} generation and update are presented focusing on those which combine \acs{fbp} with \acs{rm} generation.
\subsection{Fingerprinting based positioning}\label{subsec:fingpos}
There are two types of \acs{fbp} approaches: deterministic ones and probabilistic ones. A typical representative of the first type is \acf{knn} where the estimated position of the user is obtained as the average of the \acsp{rp} {associated with} the \acl{knn} of the measured \acs{rss} in the \acs{rss} space of the \acs{rm} \cite{Padmanabhan2000,4907834}. Probabilistic approaches \cite{Youssef2008,Ndrmyr2014} use the likelihood of the observed \acs{rss} and a proper prior distribution of the position to compute the posterior and estimate the users location, e.g. by applying the \acf{map} principle.

Here we focus on reviewing \acs{fbp} methods using \acsp{nn}. These are the typical \acs{fbp} approaches utilizing machine learning algorithms, see e.g., \cite{He2016}. In \cite{Zh2017,Xu2016}, position estimation methods based on \acsp{nn} with 1 hidden layer and nonlinear activation functions, i.e. introducing nonlinear transformation to the input of the \acs{nn}, are proposed. The positioning methods presented in above publications are deterministic, and they have been shown these to perform than \acs{knn}. Apart from position estimation based on \acs{nn} with a single hidden layer, Zhang et al. \cite{Zhang2016} employed a \acf{sdae}, which is a \acf{dnn} consisting of multiple hidden layers, to \acs{fbp} . They train \acs{sdae} model using 100 \acsp{rm} which are built from manual site surveys throughout the \acs{roi} at different times and each \acs{rm} is treated as one training example.

The \acs{svbi} based position estimation which we propose herein is also realized using \acsp{nn}. However, it differs from the previous work in two aspects. On the one hand, it is a probabilistic model for position estimation. On the other hand, it can be trained using individual observation, i.e. each \acs{rp}  with the corresponding \acs{rss} readings is used to train the \acs{nn}, instead of treating the entire \acs{rm} as only one training example.
\subsection{Joint \acs{fbp} and \acs{rm} generation}\label{subsec:jointfbprmcons}
In this part, we categorize the radio map construction approaches according to whether they combine \acs{fbp} and \acs{rm} generation. Herein generation of \acs{rm} means that {an} \acs{rm} is newly constructed from the already existing \acs{rm} and additionally measured \acs{rss} without associated position. Most of the previous publications are focused on generating the \acs{rm} without jointly estimating the position \cite{Zh2017,He2016,Talvitie2015,Atia2013}. In \cite{Talvitie2015}, Talvitie et al. apply interpolation and extrapolation of \acs{rss} based on the distances in the coordinate space to generate a \acs{rm} which includes more \acsp{rp} than the original \acs{rm}. This helps to include \acsp{rp} into the \acs{rm} which could not be occupied during site survey, e.g. because of access restrictions. The disadvantages of this method are twofold: i) the accuracy of the interpolated/extrapolated \acs{rm} depends on the number and spatial distribution of the \acsp{rp} included in the collected \acs{rm}, and ii) it needs a separate \acs{fbp} approach for position estimation.

There are few publications addressing simultaneous \acs{fbp} and \acs{rm} generation. Feng et al. \cite{Feng2012} and Majeed et al. \cite{Majeed2016} employ compressive sensing with ${L}_1 $ regularization and manifold alignment with geometry perturbation to achieve both \acs{fbp} and \acs{rm} generation. In \cite{Zh2017}, Zhou et al. propose a method using \acsp{nn} with backpropagation to realize joint position estimation and \acs{rm} generation. However, the intrinsic discrepancy of the dimensionality between \acs{rss} readings and the coordinates requires to train the position estimation and \acs{rm} generation models separately, i.e. there are two different training processes: one for mapping the \acs{rss} to the coordinate space to achieve \acs{fbp}, and another one for transforming the coordinates to \acs{rss} space for \acs{rm} generation.

The proposed approach makes use of a latent representation of the \acs{rss} readings via \acs{svbi} in order to achieve both position and \acs{rss} estimation. These two estimation processes are implemented jointly. In this way, it differs significantly from previous work.
\section{Fundamentals}\label{sec:fundamentals}
The goal of this paper is to propose a probabilistic model for achieving joint \acs{fbp} and \acs{rm} generation implemented using a \acs{dnn} unifiedly. Having referred to related work above, we now introduce the notation and mathematical concepts used {lateron}.
\subsection{Fingerprinting based positioning}\label{subsec:fundfbp}
During the offline stage, a site survey is conducted to construct the fingerprinting database, namely the \acf{rm}. It consists of collecting the \acs{rss} readings of signals from all \acsp{ap} at each \acs{rp} using a suitable device (e.g., the mobile phone) with the \acs{wifi} module. The coordinates of the \acsp{rp} must be either known beforehand or determined independent while collecting the \acs{rss}. Assuming that an \acs{rm} consisting of $ N_{RP} $ \acsp{rp} and the \acs{rss} readings from $N_{AP}$ \acsp{ap} are available at each of these \acsp{rp}, we can express the \acs{rm} as $ RM=\left[\mathbf{Y}^{N_{RP}}, \mathbf{X}^{N_{RP}}\right]\in\mathcal{R}^{N_{RP}\times(D+N_{AP})}$, where $ \mathbf{Y}^{N_{RP}} $ is the matrix of coordinates of the \acsp{rp} in $ D $ dimensions (e.g., $ D=2\, \text{or}\, 3 $), and $ \mathbf{X}^{N_{RP}} $ is the matrix of the readings of \acs{rss}. The $ i^{\text{th}} $ row in $ RM $, i.e. $ \left[\mathbf{y}_{i}^{\mathrm{T}},\mathbf{x}_{i}^{\mathrm{T}}\right] $, denotes the coordinates $\mathbf{y}_i$ of the $i^{\mathrm{th}}$ \acs{rp} and the vector $\mathbf{x}_i$ of \acs{rss} values associated with that \acs{rp}. A fingerprinting based positioning approach (e.g., \acs{knn}) can be interpreted as a mapping $ \varphi_{RM}:\mathbf{x}\mapsto\mathbf{y} $ from \acs{rss} space to coordinate space. At the online stage, $ \varphi_{RM} $ is applied to compute the current location $ \hat{\mathbf{y}}_{u} $ of the user from currently measured \acs{rss} readings $ \mathbf{x}_{u} $, i.e. $ \hat{\mathbf{y}}_u=\varphi_{RM}(\mathbf{x}_u) $.
	
To evaluate the positioning accuracy, a test dataset, which is not used to train the \acs{fbp} model, {i.e. to set up the \acs{rm} in the case of \acs{knn}}, is introduced. Similar to the notation used for the $ RM $, let test dataset consist of $ N_{TP} $ \acfp{tp} and collect the corresponding coordinates and \acs{rss} values in the matrix $ TS=\left[\mathbf{Y}^{N_{TP}}, \mathbf{X}^{N_{TP}}\right] $. The coordinates in $ TS $ are only used for performance evaluation via computing the error (e.g., \acf{mse}, \acf{rmse}, Euclidean distance) between $ \mathbf{Y}^{N_{TP}} $ and the coordinates $ \hat{\mathbf{Y}}^{N_{TP}}=\varphi_{RM}(\mathbf{X}^{N_{TP}}) $ estimated from $\mathbf{X}^{N_{TP}}$.
\subsection{Neural networks and \aclp{ae}}\label{subsec:fundnnae}
We provide a short introduction to \acsp{nn} and \acfp{ae} here to support the further analysis. More details about \acsp{nn} and \acsp{ae} can be found e.g. \cite{Demuth2014,Zh2017}, and \cite{Vincent2010} respectively. 
\subsubsection{Neural networks}\label{subsubsec:nn}
A \acf{snn} consists of $d_{\mathrm{out}}$ nodes (neurons) which transform the input vector $ \mathbf{f}_{\mathrm{in}}\in\mathcal{R}^{d_{\mathrm{in}}} $ into an output vector $ \mathbf{f}_{\mathrm{out}}\in\mathcal{R}^{d_{\mathrm{out}}} $. The elements of $\mathbf{f}_{\mathrm{in}}$ are linearly combined using weights $\mathbf{w}$ collected in a matrix $ \mathbf{W}\in\mathcal{R}^{d_\mathrm{in}\times d_\mathrm{out}}$, shifted using biases $ \mathbf{b}\in\mathcal{R}^{d_\mathrm{out}} $ and then processed by an activation function $ \mathbf{f}_\mathrm{act} $ (\figpref\ref{subfig:snn}). The free parameters of an \acs{snn} are the weights, biases and activation function. For using the \acs{snn} these free parameters need to be determined by optimization such that the error of the output
\begin{equation}
\label{eq:snnoutput}
\hat{\mathbf{f}}_{\mathrm{out}}^{\mathrm{{SNN}}}=\mathbf{f}_{\mathrm{act}}(\mathbf{W}^{\mathrm{T}}\mathbf{f}_{\mathrm{in}}+\mathbf{b})
\end{equation}
is minimized for a given training data. Generally, $ \mathbf{f}_\mathrm{act} $ is chosen, and gradient descent (e.g., \acs{bfgs}, \acl{lm}) is applied to back propagate the error (e.g., \acl{se} between $ \mathbf{f}_{\mathrm{out}} $ and the output of the \acs{snn}, i.e. $ \|\mathbf{f}_\mathrm{out}-\hat{\mathbf{f}}_{\mathrm{out}}^{\mathrm{SNN}}\|^{2}_{2} $) for training and optimizing the \acs{snn} \acs{wrt} $ \mathbf{W} $ and $ \mathbf{b} $\cite{LeCun2012}.
	
Generalizing from the \acs{snn}, we can progress to a \acf{dnn} that consists of $ L $ layers, whose corresponding configuration is given the weights, biases, and activation function for each layer, i.e. $ \{\mathbf{W}_i, \mathbf{b}_i, f_{\mathrm{act}}^{i}\}_{i=1}^{L}$ (\figpref\ref{subfig:dnn}). Herein the number of nodes in each layer is equal to the number of columns of the weight matrix of the corresponding layer. Similarly, the output of the $ L $ layers \acs{dnn} can be written as:
	\begin{equation}
		\label{eq:dnnoutput}
		\begin{split}
			\hat{\mathbf{f}}_{\mathrm{out}}^{\mathrm{DNN}} &= \underbrace{\mathbf{f}_{\mathrm{act}}^{L}(\mathbf{W}_L^{\mathrm{T}}\mathbf{f}^{L-1}_{\mathrm{act}}(\cdots\mathbf{f}^1_{\mathrm{act}}(\mathbf{W}_{1}^{\mathrm{T}}\mathbf{f}_{\mathrm{in}}+\mathbf{b}_1)\cdots)+\mathbf{b}_L)}_{f(\mathbf{f_{\mathrm{in}}})}\\
			&:= {f(\mathbf{f_{\mathrm{in}}})}
		\end{split}
	\end{equation}
	
For the subsequent analysis, we build several \acfp{bm} for \acs{fbp} using an \acs{snn} or a \acs{dnn}, i.e. the feed-in and feed-out of the \acsp{nn} are the readings of \acs{rss} and the corresponding coordinates of the \acs{rp}. We analysis the performance of the \acsp{bm} and compare to other models (See \secpref\ref{sec:expanadis}).
\subsubsection{Auto-encoders}\label{subsubsec:ae}
\begin{figure}[!htb]
		\label{fig:snndnnae}
		\centering
		\subfloat[\acs{snn}]{
			\label{subfig:snn}
			\includegraphics[width=0.5\columnwidth]{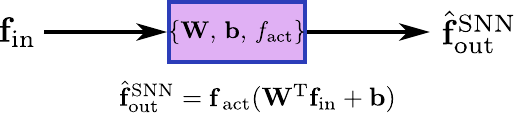}
		}\\
		\subfloat[\acs{dnn}]{
			\label{subfig:dnn}
			\includegraphics[width=0.95\columnwidth]{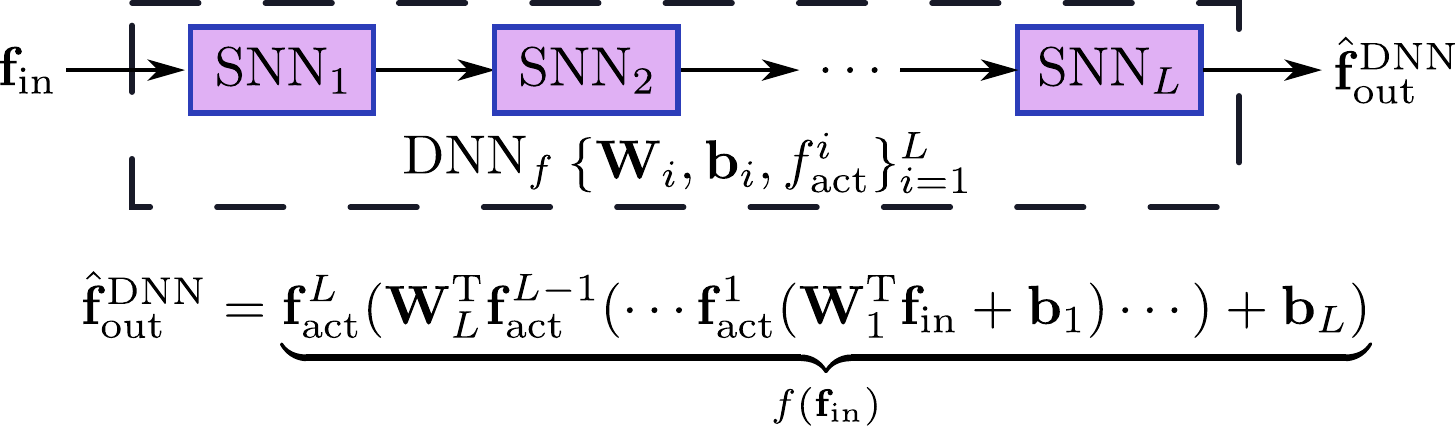}
		}\\
		\subfloat[\acs{ae}]{
			\label{subfig:ae}
			\includegraphics[width=0.75\columnwidth]{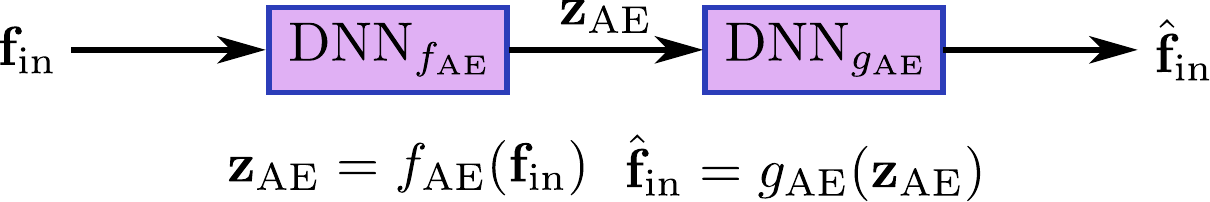}
		}
		\caption{Schematic depiction of \acs{snn}, \acs{dnn}, and \acs{ae}}			
	\end{figure}

A deep network with a denoising capacity is required instead of \acs{snn} and \acs{dnn} if the feed-in is strongly affected by noise\cite{Vincent2010}. An \acf{ae} is a self-supervised learning model whose feed-out is the same as the feed-in and it is implementable via two pipelined \acsp{dnn}, i.e. an encoder, and a decoder, as shown in \figpref\ref{subfig:ae}. \acsp{ae} and their improved versions (e.g., \acs{sdae}) are designed to mitigate the noisy inputs. The encoder transforms the input $ \mathbf{f}_{\mathrm{in}} $ into a latent representation $ \mathbf{z}_{\mathrm{AE}} $ via the first \acs{dnn}, denoted as $ f_{\mathrm{AE}} $, i.e.:
	\begin{equation}
		\label{eq:enc}
		\mathbf{z}_{\mathrm{AE}}={f}_{\mathrm{AE}}(\mathbf{f}_{\mathrm{in}})
	\end{equation}     
The decoder reconstructs $ \mathbf{f}_{\mathrm{in}} $ from the latent representation using another \acs{dnn} $ g_{\mathrm{AE}} $, i.e.:
	\begin{equation}
		\label{eq:dec}
		\hat{\mathbf{f}}_{\mathrm{in}}=g_{\mathrm{AE}}(\mathbf{z}_{\mathrm{AE}})
	\end{equation}
Techniques for training an \acs{ae} have been proposed by Lee et al. \cite{lee2008sparse}, and Hinton et al. \cite{hinton2006reducing}. They have been proved successful at training a \acs{dnn} to learn their parameters such that minimize the difference between $ \hat{\mathbf{f}}_{\mathrm{in}} $ and $ \mathbf{f}_{\mathrm{in}} $. Above brief introduction to \acs{ae} is helpful to understand the reason why \acs{svbi} based joint {position} and \acs{rss} estimation can be implemented using \acsp{nn} (See \secpref\ref{sec:svbi}). 

	\section{\acl{svbi} and its unified implementation} \label{sec:svbi}
	Stochastic variational Bayesian inference (\acs{svbi}) is applied to formulate a probabilistic model, which not only can model the posterior, i.e. $ p_{\mathbf{\theta}}(\mathbf{y}|\mathbf{x}) $ ($\mathbf{x}$ is a vector representing the \acs{rss} and $\mathbf{y}$ is the associated coordinates of the \acs{rp}), used for \acs{fbp}, but also is capable of modeling the feed-in of the positioning model, i.e. the density $ p_{\mathbf{\theta}}(\mathbf{x}) $ of the readings of \acs{rss} \footnote{Actually the density of each observation of \acs{rss} is dependent on the location where it is measured, i.e. $ p_{\mathbf{\theta}(\mathbf{y})}(\mathbf{x}) $. Herein we use a simplified notation $ p_{\mathbf{\theta}}(\mathbf{x}) $ to denote the density function of the \acs{rss}.}. $ \mathbf{\theta} $ is the parameters to determine the assumed distribution (e.g., mean and covariance of a Gaussian distribution). Firstly, the analysis on estimating $ p_{\mathbf{\theta}}(\mathbf{x}) $ using \acs{svbi}, including some fundamentals, the optimization scheme and the relationship with \acsp{ae}, is presented. Then we generalize it to estimate $ p_{\mathbf{\theta}}(\mathbf{y}|\mathbf{x}) $ as well as the joint distribution $ p_{\mathbf{\theta}}(\mathbf{x}, \mathbf{y}) $. Finally, a unified frameworks illustrating the implementation of \acs{svbi} and its application to \acs{fwips},  especially for joint \acs{fbp} and \acs{rm} generation, is analyzed       
	\subsection{\acs{svbi} based density estimation}
	\subsubsection{Fundamentals of \acs{svbi}}
	Assume that each of the $ \mathbf{x}_{i}$ collected in $\mathbf{X}^{N_{\mathrm{RP}}}$ have been drawn \acs{iid} from $ p_{\mathbf{\theta}}(\mathbf{x}) $ by the process of collecting the readings of \acs{rss}. Based on the manifold assumption that the hyper-dimensional \acs{rss} lies on an intrinsic manifold, whose dimension is much lower than that of the \acs{rss}\cite{pulkkinen2011semi}, we introduce a latent variable $ \mathbf{z}\in\mathcal{R}^{d_{man}} $, where $ {d_{man}} $ is the dimension of the intrinsic manifold, to represent the information content of $\mathbf{x}$ . The density of $ \mathbf{x} $ can be computed via marginalization over the latent space:
	\begin{equation}
		\label{eq:margin}
		p_{\mathbf{\theta}}(\mathbf{x})=\int{p_{\mathbf{\theta}}(\mathbf{x}, \mathbf{z})}d\mathbf{z}
	\end{equation}
	where $p_{\mathbf{\theta}}(\mathbf{x}, \mathbf{z})$ is the joint distribution. By introducing a mapping distribution, i.e. $ q_{\mathbf{\phi}}(\mathbf{z}|\mathbf{x}) $, where $ \mathbf{\phi} $ is the vector of parameters of the mapping distribution, \eqref{eq:margin} can be rewritten as:
	\begin{equation}
		\label{eq:is} 
		\begin{split}
			p_{\mathbf{\theta}}(\mathbf{x}) &= \int{p_{\mathbf{\theta}}(\mathbf{x}|\mathbf{z})p_{\mathbf{\theta}}(\mathbf{z})\frac{q_{\mathbf{\phi}}(\mathbf{z}|\mathbf{x})}{q_{\mathbf{\phi}}(\mathbf{z}|\mathbf{x})}}d\mathbf{z}\\
			&= \mathbb{E}_{q_{\mathbf{\phi}}(\mathbf{z}|\mathbf{x})}\left[p_{\mathbf{\theta}}(\mathbf{x}|\mathbf{z})\frac{p_{\mathbf{\theta}}(\mathbf{z})}{q_{\mathbf{\phi}}(\mathbf{z}|\mathbf{x})}\right]
		\end{split}
	\end{equation} 
	where $\mathbb{E}$ denotes th expectation operator and $p_{\mathbf{\theta}}(\mathbf{z})$ is the distribution of $\mathbf{z}$. $ q_{\mathbf{\phi}}(\mathbf{z}|\mathbf{x}) $ can be interpreted as an encoder similar to the one within an \acs{ae} and represents the distribution of the mapping from $\mathbf{x}$ to $\mathbf{z}$. The goal of \acs{svbi} is to maximize the estimated probability of $\mathbf{x}$, which is equivalent to maximizing the logarithm of the density function of it \acs{wrt} to $\mathbf{\theta}$. Taking the logarithm of \eqref{eq:is} and applying Jensen's Inequality \footnote{Jensen's Inequality gives the lower bound of convex functions. Recall that $g(x)$ is a convex function if, for $0<\lambda<1$, $  g(\lambda x + (1-\lambda)y) \leq \lambda{g(x)} + (1-\lambda)g(y)$ for all $x$ and $y$. Conversely, $g(x)$ is concave if $-g(x)$ is convex.}  to it, the lower bound of \eqref{eq:is} is:
	\begin{equation}
		\label{eq:lowerbound}
		\begin{split}
			\ln{p_{\mathbf{\theta}}(\mathbf{x})}&\geq\mathbb{E}_{q_{\mathbf{\phi}}(\mathbf{z}|\mathbf{x})}\left[\ln{\Big(p_{\mathbf{\theta}}(\mathbf{x}|\mathbf{z})\frac{p_{\mathbf{\theta}}(\mathbf{z})}{q_{\mathbf{\phi}}(\mathbf{z}|\mathbf{x})}\Big)}\right]\\
			&=\mathbb{E}_{q_{\mathbf{\phi}}(\mathbf{z}|\mathbf{x})}\left[\ln{p_{\mathbf{\theta}}(\mathbf{x}|\mathbf{z})}\right] - \mathbb{E}_{q_{\mathbf{\phi}}(\mathbf{z}|\mathbf{x})}\left[\ln{\frac{q_{\mathbf{\phi}}(\mathbf{z}|\mathbf{x})}{p_{\mathbf{\theta}}(\mathbf{z})}}\right]
		\end{split}		
	\end{equation}
	The second term on the \acf{rhs} of \eqref{eq:lowerbound} is Kullback-Leibler divergence between $q_{\mathbf{\phi}}(\mathbf{z}|\mathbf{x})$ and $p_{\mathbf{\theta}}(\mathbf{z})$, denoted as $ \mathrm{D}_{\mathrm{KL}}(q_{\mathbf{\phi}}(\mathbf{z}|\mathbf{x})\|p_{\mathbf{\theta}}(\mathbf{z})) $ and introducing $ \mathcal{L}(\mathbf{\theta}, \mathbf{\phi}) $ for the variational lower bound of the marginalized likelihood, we obtain:
	\begin{equation}
		\label{eq:objlowerbound}
		\mathcal{L}(\mathbf{\theta}, \mathbf{\phi}) = \mathbb{E}_{q_{\mathbf{\phi}}(\mathbf{z}|\mathbf{x})}\left[\ln{p_{\mathbf{\theta}}(\mathbf{x}|\mathbf{z})}\right]- \mathrm{D}_{\mathrm{KL}}(q_{\mathbf{\phi}}(\mathbf{z}|\mathbf{x})\|p_{\mathbf{\theta}}(\mathbf{z}))
	\end{equation}
	The maximization of $ \mathcal{L}(\mathbf{\theta}, \mathbf{\phi}) $ \acs{wrt} $ \mathbf{\theta} $ and $ \mathbf{\phi} $ is equivalent to minimizing the second term of \acs{rhs} of \eqref{eq:objlowerbound}, because Kullback-Leibler divergence is non-negative. Often, mean-field approach is applied to approximate $ \mathbf{\theta} $ and $ \mathbf{\phi} $ \cite{Kingma2013}. However, this requires the analytical solutions of the expectations\cite{Kingma2013}. In fact, the analytical solutions of the expectation is unknown for the most cases.
	
	Instead of maximizing using mean-field approach by computing the analytical solution of the expectation, we use gradient based backpropagation to optimize $ \mathcal{L}(\mathbf{\theta}, \mathcal{\phi}) $ \acs{wrt} $ \mathbf{\theta}, \mathbf{\phi} $, i.e. step-wisely searching the optimal values of $ \mathbf{\theta}, \mathcal{\phi} $. The step size is $ \Delta_{\mathbf{\theta}, \mathbf{\phi}} $:
	\begin{equation}
		\label{eq:rmsprop}
		\Delta_{\mathbf{\theta}, \mathbf{\phi}}=-\Gamma^{\mathbf{\theta}, \mathbf{\phi}}\nabla_{\mathbf{\theta}, \mathcal{\phi}}{\mathcal{L}(\mathbf{\theta}, \mathbf{\phi})}
	\end{equation}
	where $ \Gamma^{\mathbf{\theta}, \mathbf{\phi}} $ is a diagonal matrix containing the adaptive learning rate. Herein \acs{rmsprop} is used to compute it\cite{Kingma2015}. However, the indirect dependency on $\mathbf{\phi} $ over which the expectation of the first term on \acs{rhs} of \eqref{eq:objlowerbound} is taken, makes it difficult to compute the gradient of the expectation \acs{wrt} $\mathbf{\phi}$. To estimate the gradient of the expected reconstruction loss, i.e. the difference between the reconstructed $\mathbf{x}$ from $\mathbf{z}$ and the measured $\mathbf{x}$, \acs{wrt} $ \mathbf{\phi} $, Schulman et al. proposed stochastic computation graphs\footnote{This approach is also named pairwise derivative, infinitesimal perturbation analysis, and stochastic backpropagation in other publications.} to calculate the gradient\cite{Schulman2015,Paisley2012}. This approach solves the indirect dependency problem by applying re-parameterization trick. More details can be found in e.g. \cite{Kingma2013,Rezende2014,Paisley2012,Devroye1986}. 
	
	The re-parameterization trick is formulated in the following way. Assume that the latent variable $ \mathbf{z} $ can be expressed by another deterministic variable $ \mathbf{z}=\gamma_{\mathbf{\phi}}(\mathbf{\epsilon}, \mathbf{x}) $, where $ \mathbf{\epsilon} $ is an auxiliary variable distributed corresponding to an independent distribution, i.e. $ \mathbf{\epsilon}\sim p(\mathbf{\epsilon}) $, and $ p(\mathbf{\epsilon}) $ is an arbitrary distribution (e.g., Gaussian). Under this assumption, the expected reconstruction loss, i.e. the \acs{rhs} of \eqref{eq:lowerbound}, is independent of the dependency on $ \mathbf{\phi} $ while taking the expectation over $ \mathbf{z} $ and can be estimated by:
	\begin{equation}
		\label{eq:noklsgvb}
		\begin{split}
			&\hat{\mathcal{L}}_1(\mathbf{\theta}, \mathbf{\phi})=\frac{1}{N_{\mathrm{MCS}}}\sum_{l=1}^{N_{\mathrm{MCS}}}\left[\ln p_{\mathbf{\theta}}(\mathbf{x},\mathbf{z}^{l})-\ln q_{\mathbf{\phi}}(\mathbf{z}^{l}|\mathbf{x})\right]\\
			&\mathrm{where}\,\mathbf{z}^{l}=\gamma_{\mathbf{\phi}}(\mathbf{\epsilon}^{l},\mathbf{x}), \mathbf{\epsilon}^{l}\sim p(\mathbf{\epsilon}), l=1,2,\cdots, N_{\mathrm{MCS}}
		\end{split}
	\end{equation}
	In \eqref{eq:noklsgvb}, $ N_{\mathrm{MCS}} $ is the number of \acf{mcs}. In the case of the Kullback-Leibler divergence can be computed analytically, another equation for evaluating \eqref{eq:lowerbound} is:
	\begin{equation}
		\label{eq:klsgvb}
		\begin{split}
			\hat{\mathcal{L}}_2&(\mathbf{\theta}, \mathbf{\phi})=\\
			&-\mathrm{D}_{\mathrm{KL}}(q_{\mathbf{\phi}}(\mathbf{z}|\mathbf{x})\|p_{\mathbf{\theta}}(\mathbf{z}))+\frac{1}{N_{\mathrm{MCS}}}\sum_{l=1}^{N_{\mathrm{MCS}}}p_\mathbf{\theta}(\mathbf{x}|\mathbf{z}^{l})
		\end{split}		
	\end{equation}
	Both \eqref{eq:noklsgvb} and \eqref{eq:klsgvb} are treated as the surrogate losses (i.e. substitute losses) of the variational lower bound. But the latter surrogation typically has lower variance than the former\cite{Kingma2013}.  With the surrogate loss, it is easier to compute the gradient of the expected loss. Therefore, stochastic gradient descent approaches (e.g., \acs{rmsprop} or \acs{adam}) can be applied to compute the optimal values of $\mathbf{\phi}$ and $\mathbf{\theta}$.
	
	\subsubsection{\acs{ae} interpretation}
	We can refer to $ q_{\mathbf{\phi}}(\mathbf{z}|\mathbf{x}) $ as the encoder, which estimates the distribution over all $ \mathbf{z} $ given an observation $ \mathbf{x} $. From the latent variable, $ \mathbf{x} $ can be reconstructed via sampling from $ p_{\mathbf{\theta}}(\mathbf{x}|\mathbf{z})$, which is equivalent to the decoding process of the \acs{ae}. Different from the \acsp{ae} described in \secpref\ref{subsec:fundnnae}, both the encoding and decoding procedures of \acs{svbi} are probabilistic.
	\subsubsection{\acs{svbi} under Gaussian assumptions}
	Herein we assume that there exists a transformation that transforms $ \mathbf{x}\sim p_{\mathbf{\theta}}(\mathbf{x}) $ to a latent variable $ \mathbf{z} $, which is standard Normally distributed, i.e. $ \mathbf{z}\sim p_{\mathbf{\theta}}(\mathbf{z}),\, p_{\mathbf{\theta}}(\mathbf{z})=\mathcal{N}(\mathbf{z}|0,\mathrm{I})$. A \acs{dnn} is employed to approximate this transformation. The variational posterior $ q_{\mathbf{\phi}}(\mathbf{z}|\mathbf{x}) $, i.e. the encoder, is assumed to be a multivariate Gaussian:
	\begin{equation}
		\label{eq:qgauss}
		 q_{\mathbf{\phi}}(\mathbf{z}|\mathbf{x}) = \mathcal{N}(\mathbf{z}|\mathbf{\mu}_{\mathbf{z}},\Sigma_{\mathbf{z}})
	\end{equation}
	where $ \Sigma_{\mathbf{z}} $ can be written as $ \Sigma_{\mathbf{z}}:=\mathbf{R}\mathbf{R}^{\mathrm{T}},\mathbf{R}\in\mathcal{R}^{d_{\mathrm{man}}\times d_{\mathrm{man}}} $, $ \mathbf{R} $ is the Cholesky decomposition of $ \Sigma_{\mathbf{z}} $. This decomposition is to reduce the number of parameters needed be estimated. In this case, $ \mathbf{z} $ can be re-parametrized by a linear transformation such that $ \mathbf{z}=\mathbf{\mu}_{\mathbf{z}}+\Sigma_{\mathbf{z}}^{1/2}\mathbf{\epsilon}, \mathbf{z}\sim \mathcal{N}(\mathbf{z}|\mathbf{\mu}_{\mathbf{z}},\Sigma_{\mathbf{z}})$, where  $ \mathbf{\epsilon}\sim \mathcal{N}(0,\mathrm{I}), \mathbf{\epsilon}\in \mathcal{R}^{d_{\mathrm{man}}} $. Therefore, the Kullback-Leibler divergence in \eqref{eq:klsgvb} can be analytically written as \eqref{eq:anakl} under above assumptions \cite{IMM2012-03274}.
	\begin{equation}
		\label{eq:anakl}
		\begin{split}
			\mathrm{D}_{\mathrm{KL}}&(q_{\mathbf{\phi}}(\mathbf{z}|\mathbf{x})\|p_{\mathbf{\theta}}(\mathbf{z}))=\\
			&-\frac{1}{2}\left[d_{\mathrm{man}}+\ln|\Sigma_{\mathbf{z}}|-\mathrm{Tr}(\Sigma_{\mathbf{z}})-\mathbf{\mu}_{\mathbf{z}}^{\mathrm{T}}\mathbf{\mu}_{\mathbf{z}}\right]
		\end{split}	
	\end{equation}
	Instead of directly estimating the distribution of $ \mathbf{x} $, we use \acf{mle} to estimate the values of $ \mathbf{x} $ with the assumed intrinsic distribution of $ p_{\mathbf{\theta}} (\mathbf{x})$, i.e. $ \hat{\mathbf{x}}=\mathrm{MLE}(\mathbf{x}), \mathbf{x}\sim p_{\mathbf{\theta}}(\mathbf{x}|\mathbf{z})$. The \acs{mle} can be implementable using \acsp{nn}\cite{Bishop:2006:PRM:1162264}. Herein we assume that $ p_{\mathbf{\theta}} (\mathbf{x})$ is Gaussian distributed, \acs{mle} can be realized by a 2 layers \acs{nn}, whose activation functions are tangent hyperbole and linear respectively\cite{Kingma2013,Rezende2014}. Based on the \acs{mle}, we introduce another surrogate loss of \eqref{eq:klsgvb} given the feed-in dataset $ \mathbf{X}^{N_{\mathrm{RP}}} $:
	\begin{equation}
		\label{eq:mseloss}
		\begin{split}
			\tilde{\mathcal{L}}_{2}^{\mathrm{MSE}}&(\mathbf{\theta},\mathbf{\phi})=-\frac{1}{N_{\mathrm{RP}}}\sum_{i=1}^{N_{\mathrm{RP}}}\mathrm{D}_{\mathrm{KL}}(q_{\mathbf{\phi}}(\mathbf{z}|\mathbf{x}_{i})\|p_{\mathbf{\theta}}(\mathbf{z}))\\
			&  + \frac{1}{N_{\mathrm{RP}}}\left[\sum_{i=1}^{N_{\mathrm{RP}}}{\|\mathbf{x}_i-\hat{\mathbf{x}}_i\|_2^2}\right]\\
			&\mathrm{where}\,\, \hat{\mathbf{x}}_i=\mathrm{MLE}(p_{\mathbf{\theta}}({\mathbf{x}}_i|\mathbf{z})), \forall \mathbf{x}_i \in \mathbf{X}^{N_{\mathrm{RP}}}
		\end{split}
	\end{equation}
	\subsection{Generalization of \acs{svbi}}
	In the case of \acf{fbp}, we try to use \acs{svbi} to maximize the conditional likelihood $ p_{\mathbf{\theta}}(\mathbf{y}|\mathbf{x}),\left[\mathbf{x},\mathbf{y}\right]\in RM $. The variational lower bound of it can be directly extended from \eqref{eq:lowerbound}:
	\begin{equation}
		\label{eq:condvlb}
		\begin{split}
			\mathcal{L}^{\mathrm{cond}}(\mathbf{\theta},\mathbf{\phi})=&\mathbb{E}_{q_{\mathbf{\phi}}(\mathbf{z}|\mathbf{x},\mathbf{y})}\left[\ln p_{\mathbf{\theta}}(\mathbf{y}|\mathbf{x},\mathbf{z})\right]\\
			&-\mathrm{D}_{\mathrm{KL}}(q_{\mathbf{\phi}}(\mathbf{z}|\mathbf{x},\mathbf{y})\|p_{\mathbf{\theta}}(\mathbf{z}|\mathbf{x}))
		\end{split}
	\end{equation}
	
	In \cite{Sohn2015}, Sohn et al. relax the dependency of $\mathbf{z}  $ on $ \mathbf{x} $ in \eqref{eq:condvlb} to be statistically independent from it, i.e. $ p_{\mathbf{\theta}}(\mathbf{z}|\mathbf{x})\simeq p_{\mathbf{\theta}}(\mathbf{z}) $. This relaxation makes it easier to be implemented. \acs{svbi} can also be generalized to estimate the joint distribution $ p_{\mathbf{\theta}}(\mathbf{x},\mathbf{y}) $ and its variational lower bound is\cite{Kingma2014}:
	\begin{equation}
		\label{eq:jointvlb}
		\begin{split}
			\mathcal{L}^{\mathrm{joint}}(\mathbf{\theta},\mathbf{\phi})=&\mathbb{E}_{q_{\mathbf{\phi}}(\mathbf{z}|\mathbf{x},\mathbf{y})}\left[\ln p_{\mathbf{\theta}}(\mathbf{y},\mathbf{x}|\mathbf{z})\right]\\
			&-\mathrm{D}_{\mathrm{KL}}(q_{\mathbf{\phi}}(\mathbf{z}|\mathbf{x},\mathbf{y})\|p_{\mathbf{\theta}}(\mathbf{z}))
		\end{split}
	\end{equation}
	\subsection{\acs{nn} based implementation of \acs{svbi} and its application to \acs{fwips}}              
	\subsubsection{General implementation of \acs{svbi}}
\begin{figure}[!htb]		
		\includegraphics[width=.95\columnwidth]{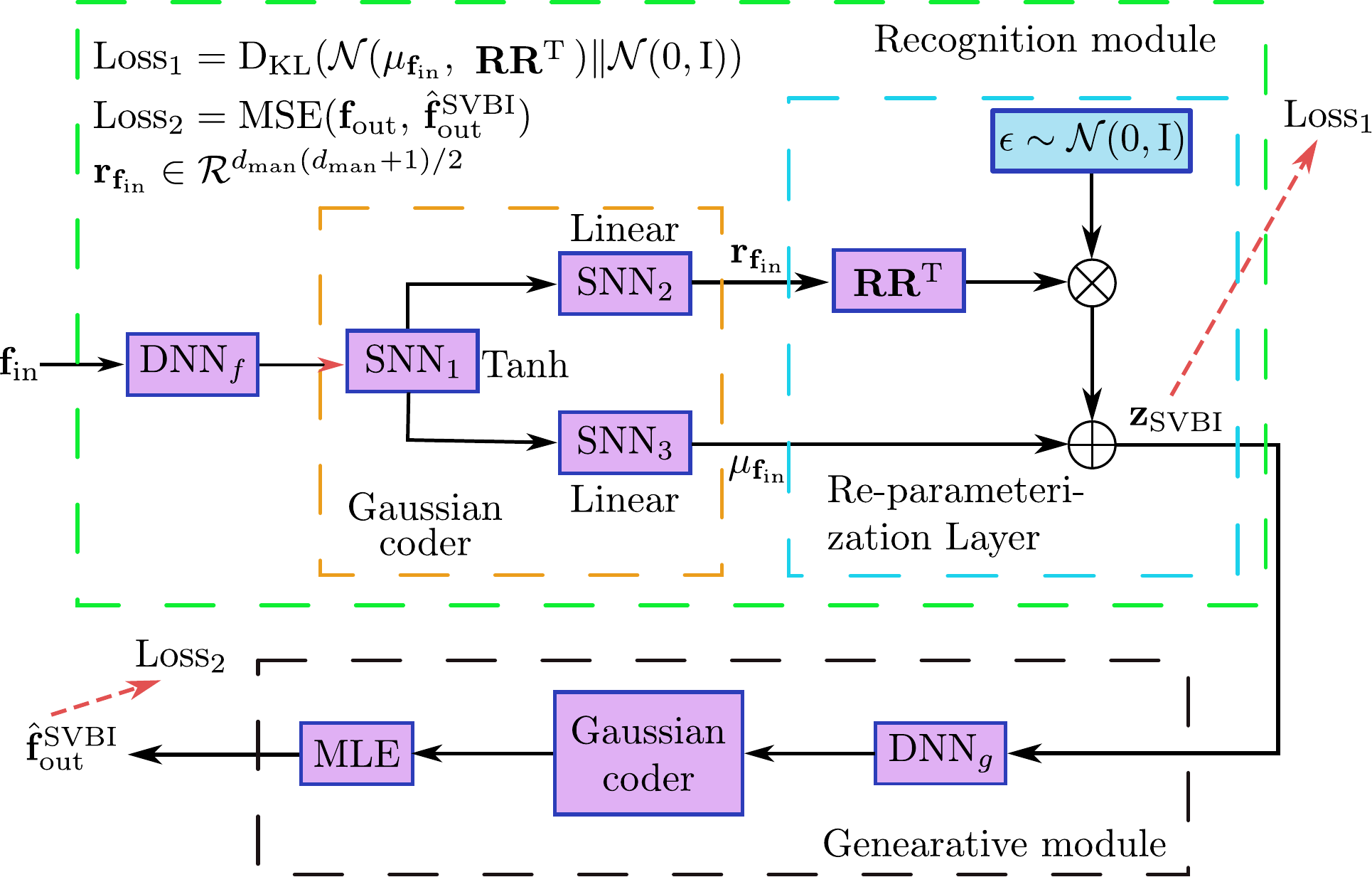}
		\caption{General implementation framework of \acs{svbi}}
		\label{fig:generalsvbi}
	\end{figure}

	First, we illustrate a general implementation framework of \acs{svbi} based on \acsp{nn}. As shown in \figpref\ref{fig:generalsvbi}\footnote{To transform $ \mathbf{r}_{\mathbf{f}_{\mathrm{in}}} $ to $\mathbf{R}$, it is reshaped to a lower triangular matrix.}, the sum of $ \mathrm{Loss}_1 $ and $ \mathrm{Loss}_2 $ is the estimated value of the variational lower bound. The scenario consists of two modules: recognition module and generative module. The former module takes $ \mathbf{f}_{\mathrm{in}} $ as the input and yields the manifold representation $ \mathbf{z}_{\mathrm{SVBI}} $ of the input. The generative module is fed by the manifold representation and trained to approximate the corresponding feed-out. Both of the modules can be implemented using \acsp{nn} with different activation functions \cite{Kingma2013}.

	This general implementation framework can be applied to estimate 3 types variational lower bound via changing the feed-in and feed-out: i) if $ \mathbf{f}_{\mathrm{in}} $ and $ \mathbf{f}_{\mathrm{out}} $ are equal to $ \mathbf{x} $, the loss is the estimation of \eqref{eq:mseloss}; ii) if $ \mathbf{f}_{\mathrm{in}} $ and $ \mathbf{f}_{\mathrm{out}} $ are a concatenation of $ \mathbf{x} $ and $ \mathbf{y} $, and $ \mathbf{y} $ respectively, the loss is an approximation of conditional variational lower bound \eqref{eq:condvlb}; and iii) if $ \mathbf{f}_{\mathrm{in}} $ and $ \mathbf{f}_{\mathrm{out}} $ are both concatenation of $ \mathbf{x} $ and $ \mathbf{y} $, it is the approximation of the joint variational lower bound \eqref{eq:jointvlb}.
	\begin{figure}[!htb]		
		\includegraphics[width=.9\columnwidth]{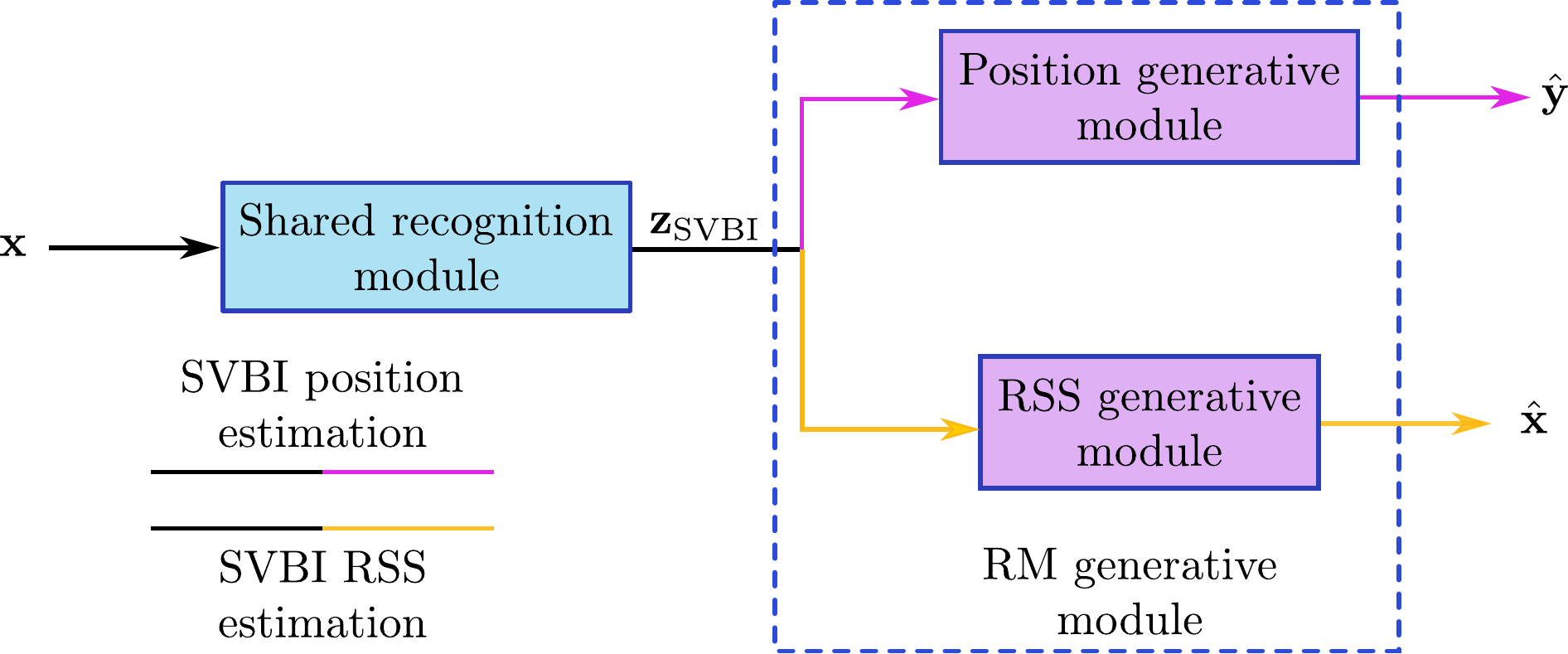}
		\caption{Hybrid \acs{svbi} based \acs{rss} and position estimation}
		\label{fig:jointsvbi}
	\end{figure}
\subsubsection{\acs{svbi} based \acs{fbp} and \acs{rss} estimation}
We now focus on applying \acs{svbi} based density estimation model jointly with the \acs{nn} based positioning approach to realize both position and \acs{rss} estimation as illustrated in \figpref\ref{fig:jointsvbi}. This joint model consists of two paths: i) the position estimation path, and ii) the \acs{rss} estimation path. Both of them has the shared recognition module but each has its own generative module for their respective purpose. In the case that there is a pre-collected \acs{rm} consisting of \acsp{rp} with associated \acs{rss} values, we can train the joint model in two ways: i) separate training, and ii) joint training. In the former case, we train the \acs{svbi} base \acs{rss} and position estimation paths separately and use it for the purpose of \acs{rss} and position estimation, respectively. As for the latter case, both paths are trained simultaneously. This joint \acs{svbi} model can be used for both position and \acs{rss} estimation, such that it can also be employed to generate the \acs{rm} by feeding the trained generative model with the arbitrary variable $ \hat{\mathbf{z}}=\hat{\mathbf{\mu}}_{\mathbf{z}}+\hat{\Sigma}_{\mathbf{z}}^{1/2}\mathbf{\epsilon}$, where $\mathbf{\epsilon}\in\mathcal{R}^{d_{\mathrm{man}}}$ is drawn from $ \mathcal{N}(0, \mathrm{I}) $.

	
In this paper, we use Keras, a deep learning library based on Tensorflow \cite{tensorflow2015-whitepaper}, to implement all the models and the simulations.
\section{Experimental Analysis and Discussion}\label{sec:expanadis}
\subsection{Testbed}
To evaluate the performance of the proposed approaches, an \acs{fwips} is deployed in the building where the Institute of Geodesy and Photogrammetry, ETH Z\"urich is located \cite{Gu2017}. This \acs{fwips} requires no additional installation of new \acsp{ap} because enough signals are available throughout the \acs{roi}. These \acsp{ap} are deployed for the purpose to provide the Internet access services. Though we cannot control the configurations of the \acs{wlan} and \acsp{ap}, we assume that the settings of them are stable. A site survey was conducted throughout the \acs{roi} using a mobile phone, Nexus 6P, with a custom made Android application in order to collect a \acs{rm}. The \acs{rp} position are determined during the site survey using a \acf{ts}, Leica MS60.  
	
Fifteen prisms permanently mounted on the ceiling represent the coordinates frame and allow the \acs{ts} to determine its position any where within the \acs{roi} by resection. A custom made frame holding the smart phone and a 360$\degree$ mini prism allows the \acs{ts} to automatically track the smart phone and thus synchronously measuring \acs{rss} and position. The position data collected by \acs{ts} have an accuracy on the {mm-level} and are thus considered as ground truth during the later analysis.

\subsection{Simulation results of the models without variational inference}	
\begin{figure}[!htb]
		\centering			
		\subfloat[\acs{bm} with post-processed scaled output]{
			\label{subfig:postscaledbm}
			\includegraphics[width=0.6\columnwidth]{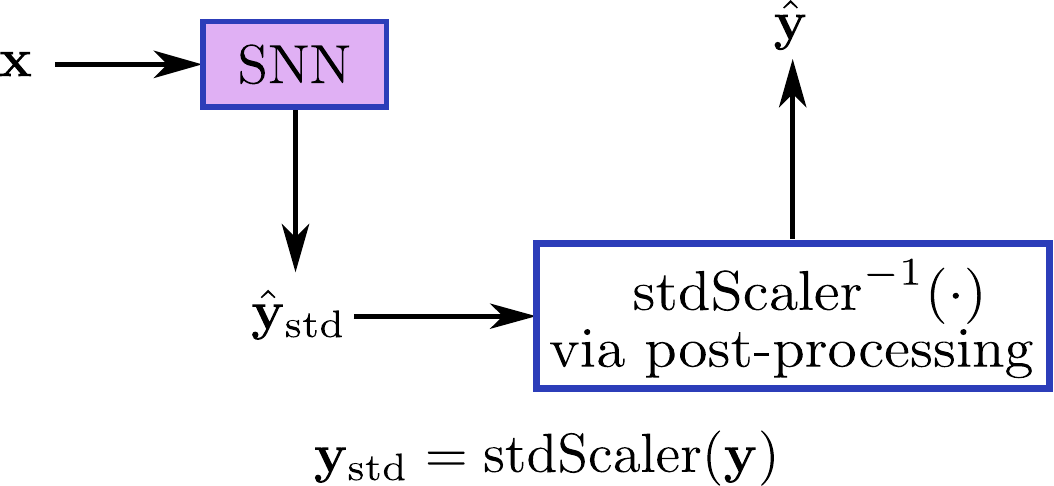}}\\
		\subfloat[\acs{bm} with built-in scaled output]{
			\label{subfig:buildinscaledbm}
			\includegraphics[width=0.6\columnwidth]{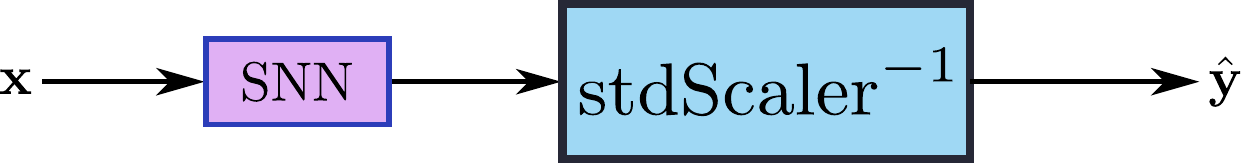}}\\
		\subfloat[\acs{dlpm}]{
			\label{subfig:dnnfip}
			\includegraphics[width=0.55\columnwidth]{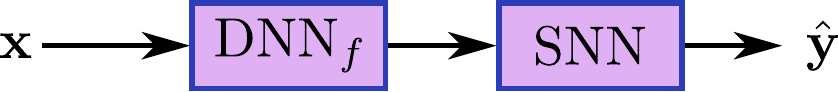}}\\
		\caption{Three \aclp{bm}}
		\label{fig:bms}
	\end{figure}

Three \acfp{bm}, illustrated in \figpref\ref{fig:bms}, for comparison are presented and analyzed, including the training and parameter setups of the \acsp{nn}, and the evaluation criterion. Instead of training the \acsp{nn} with the original \acs{rss} readings and coordinates, we start with pre-processing of the data. i) The \acs{rss} value are normalized to the range of $ \left[0, 1\right]^{N{_{\mathrm{AP}}}} $. $\mathbf{x}$ is therefore a vector of normalized values; and ii) the coordinates are standardized to have zero mean and unit standard deviation. We denote this process $ \mathrm{stdScaler}() $ and the result $\mathbf{y}_{\mathrm{std}}$ \cite{LeCun2012}. An example of standardized \acs{rm} is shown in \figpref\ref{subfig:stdorgrm}. Regarding the \acsp{bm}, the first two \acsp{bm}, shown in \figpref\ref{subfig:postscaledbm} and \figpref\ref{subfig:buildinscaledbm} respectively, are based on a \acs{snn}, whose activation function are chosen linear. The difference between the first two \acsp{bm} is the way of inverse transformation ($ \mathrm{stdScaler}^{-1}() $). The former transforms the output of the \acs{snn} to the original coordinate space after output from the \acs{snn}, and the latter realizes the inverse transformation directly within the \acs{nn}. The third \acs{bm} is a \acf{dlpm} consisting of a $L$ layers \acs{dnn} pipelined with the \acs{snn}. Thus \acs{dlpm} is a $L+1$ layers \acs{nn}. The activation functions of the  $L$ layers \acs{dnn} and \acs{snn} are set as \acf{relu}\footnote{\acs{relu} is defined as $ \mathrm{ReLu}(a)= a $ if $ a\geq 0 $ otherwise $ \mathrm{ReLu}(a)=0 $, $ a\in\mathcal{R}. $} and linear, respectively. Because we implement all the models using Keras and Tensorflow, there are two more parameters: i) batch size to fit the model, and ii) maximum number of failures of the early stopper, denoted as \acf{poe}. They are introduced to improve the training efficiency and avoid the over-fitting\footnote{We use Xavier approach \cite{Glorot2010} to initialize the weights and the biases for the \acs{nn}, and \acf{adam} and \acf{rmsprop}\cite{Kingma2015} to optimize the weights and the biases via backpropagation.}. To evaluate and compare the performance of the models, we calculate the \acs{rmse} from the Euclidean distances between the estimated coordinates $\hat{\mathbf{y}}$ and their ground truth $\mathbf{y}$ together with the 95\% confidence interval\cite{Peters2001}\footnote{If the number of trials is larger than 30, the empirical standard deviation is practically useful \cite{Peters2001}.}.
	\begin{table}[!htb]
		\caption{Basic results of the \acsp{bm}}
		\label{tab:resultsofbm}
		\centering
		\begin{tabular}{c|ccc}
			\hline
			&\acs{bm} (post)& \acs{bm} (built-in) &\acs{dlpm}\\
			\hline
			RMSE ($ \mathrm{m} $)& 2.922$ \pm 0.012$ &3.043$ \pm 0.014$ &1.634$ \pm $0.024\\
			\hline
			\hline
			&\acs{svbi} (Sep.)&SVBI (Joint)& 1-$k\mathrm{NN}$\\
			\hline
			RMSE ($\mathrm{m}$) & 1.725$ \pm 0.044$ &1.622$ \pm 0.016$&2.622\\
			\hline
		\end{tabular}
	\end{table}

The results are shown in \tabpref\ref{tab:resultsofbm}. We evaluated each model 36 times with the same configuration of the \acsp{nn} and the same training data. Since the \acsp{nn} are randomly initialized each of the 36 evaluation yielded different parameter values of the \acsp{nn} and thus slightly different results. The batch size and \acs{poe} of the \acsp{bm} were chosen as 50 and 25, respectively. The \acs{dnn} consists of 3 layers, whose number of nodes is 128, 64, 32 for the 3 layers, respectively. The performance of the \acsp{bm} with built-in and separate inverse standardization is comparable. The performance of the latter is slightly better than that of the former. \acs{dlpm} performs approximately 50\% better than that of the \acs{bm} with post inverse standardization with the data used herein. The reduction of \acs{rmse} of \acs{dlpm} is almost 0.9 $\mathrm{m}$ comparing to that of \acs{knn} \footnote{The \acs{knn} algorithm used herein is from Scikit-learn, an open source Python package for machine learning \cite{scikit-learn}. We use the weighted version of it. With our training \acs{rm}, the optimal number nearest neighbors is 1.}.
\subsection{Results with variational inference}
The \acs{dnn} of the shared recognition module is set up like the \acs{dlpm} used in the previous subsection. The \acs{rss} are mapped to a 4 dimensional latent space \footnote{We choose the dimension of latent space as 4 by trials. In the future work, we would find the optimal dimension of the latent space by cross validation.} (i.e. $d_{\mathrm{man}}=4$). To simplify the model, the number of nodes of the Gaussian coder (depicted in \figpref\ref{fig:generalsvbi}) is the same as the dimension of the latent space. Herein we present the results under the assumption that the covariance matrix of $q_\mathbf{\phi}(\mathbf{z}|\mathbf{x})$ is diagonal to reduce the computational burden.
	\begin{figure}[!htb]		
		\includegraphics[width=.9\columnwidth]{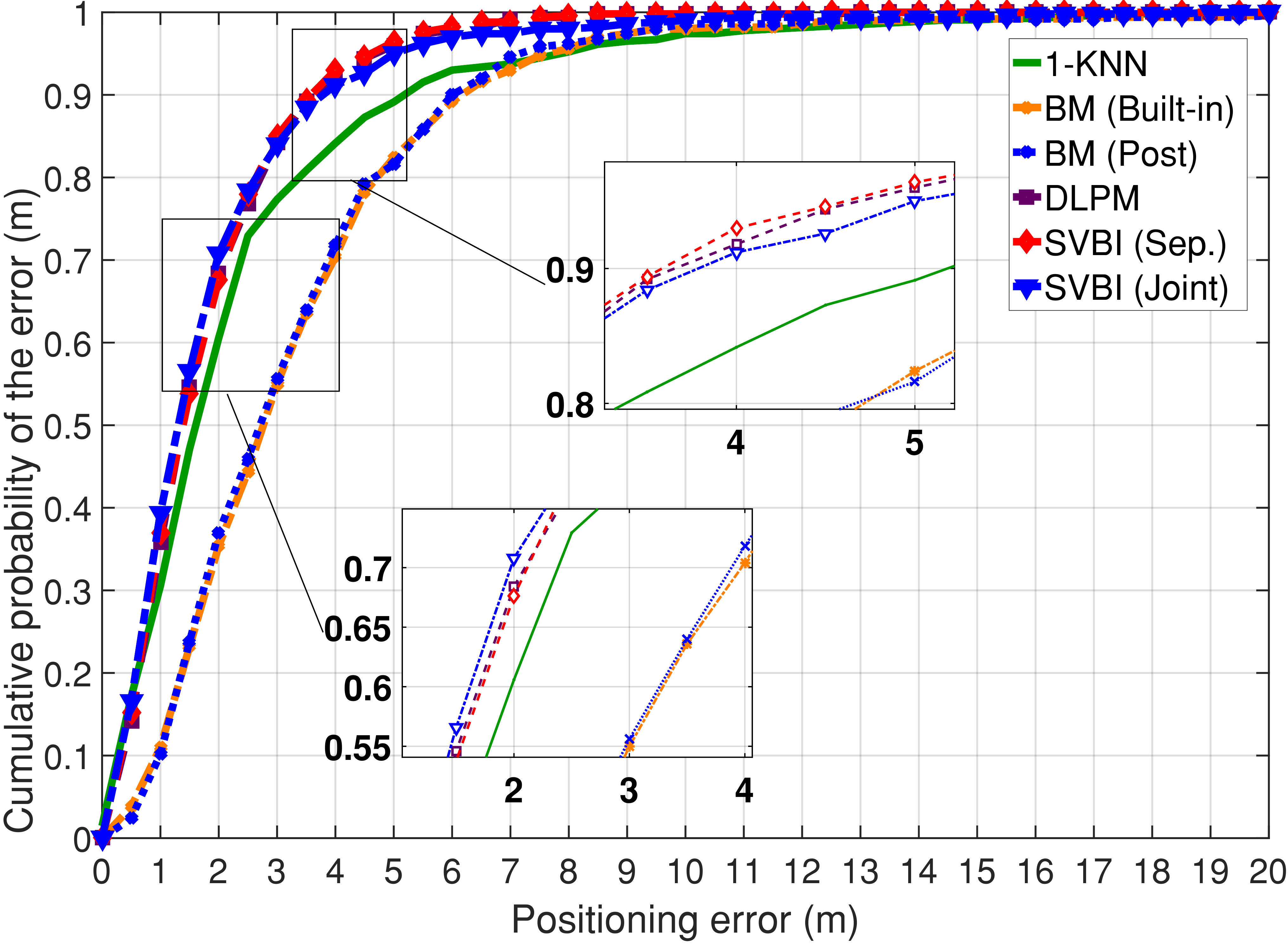}
		\caption{Cumulative positioning accuracy}
		\label{fig:cmpsvbipos}
	\end{figure}
\subsubsection{\acs{svbi} based position estimation}
To achieve \acs{svbi} based position estimation, we pipeline the shared recognition module with the \acs{bm} with the post inverse standardization. \acs{rmse} of the estimated positions then is 1.725 $\mathrm{m}$ (second part of \tabpref\ref{tab:resultsofbm} (\acs{svbi} (Sep.))). It is 40\% lower than that of \acs{bm}(post). Compared to \acs{dlpm}, \acs{rmse} of \acs{svbi} (Sep.) is slightly higher, however, from \figpref\ref{fig:cmpsvbipos}, the \acf{cpa}, defined as the cumulative density function of the positioning errors, for both models are almost the same until 2 $\mathrm{m}$.
\subsubsection{\acs{svbi} based \acs{rm} generation}
To train \acsp{nn} for the purpose of \acs{rm} generation, joint training of both paths (\figpref\ref{fig:jointsvbi}) is required. The configuration of the joint training is as follows: i) the set up of the shared recognition module and the position generative module are identical; ii) the \acs{dnn} for \acs{rss} generative module consists also of 3 layers, but with 32, 64 and 128 nodes, respectively; iii) the number of nodes of the Gaussian coder in the \acs{rss} generative module is equal to the number of \acsp{ap} (it is 399 herein); and iv) to train both paths simultaneously, we use the weighted sum of the loss of both paths.
	
We present the results of joint training in \tabpref\ref{tab:resultsofbm} (\acs{svbi}(Joint)) and \figpref\ref{fig:cmpsvbipos}. The \acs{rmse} of the positioning error of \acs{svbi}(Joint) is slightly lower than that of \acs{dlpm}. Also the \acs{cpa} of it within 2.5 $\mathrm{m}$ is 2\% higher than that of \acs{dlpm}.
		
\begin{figure}[!htb]		
		\includegraphics[width=.9\columnwidth]{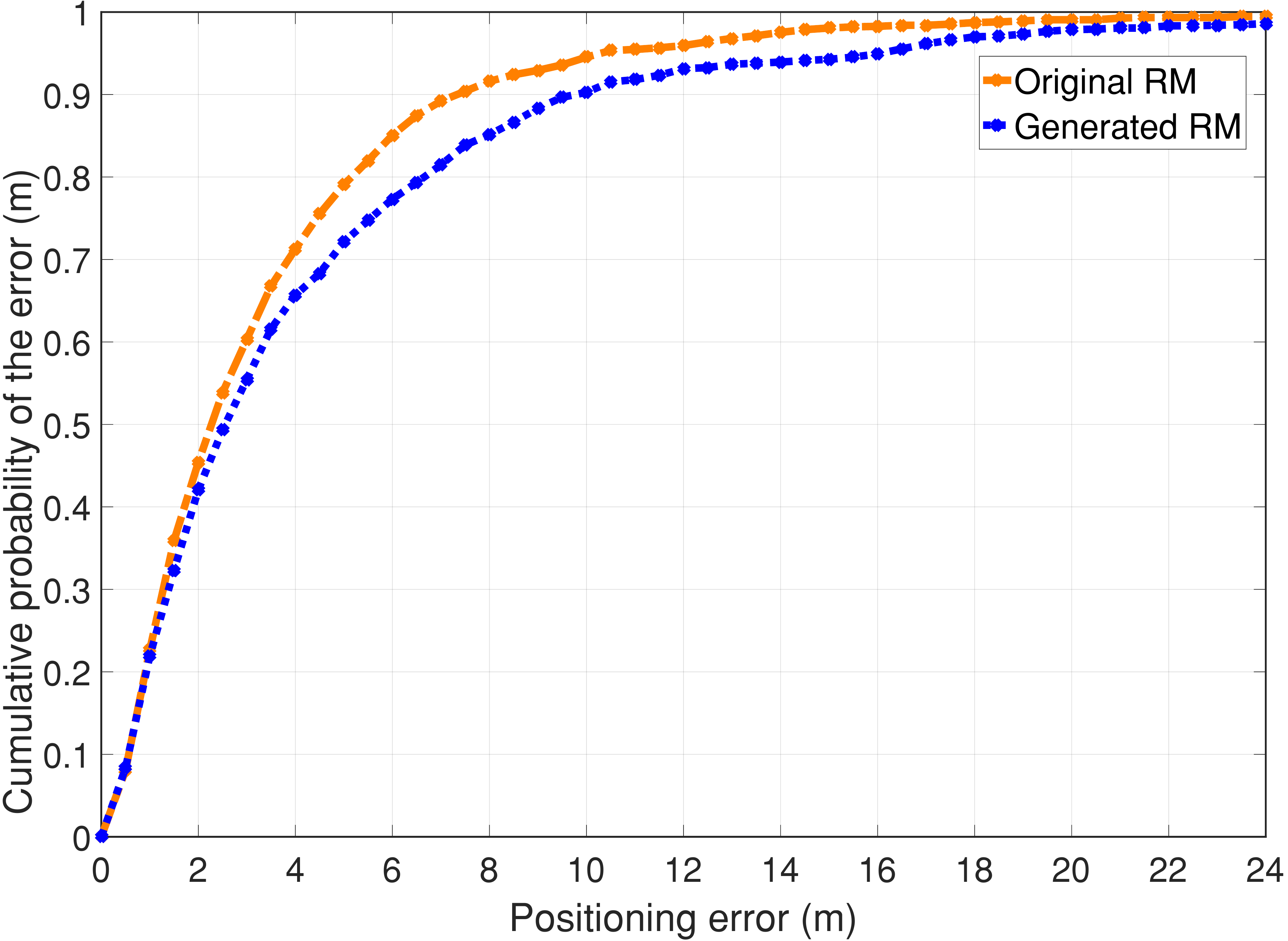}
		\caption{Comparison of \acl{cpa} using \acs{knn}}
		\label{fig:cmporgrecons}
	\end{figure}

The result of positioning error analysis of the generated \acs{rm} is illustrated in \figpref\ref{fig:cmporgrecons}. To compare the generated \acs{rm} to the original \acs{rm}, we evaluate independently using the same test dataset with \acs{knn} ($k=3$, \figpref\ref{subfig:knngraph}). The \acs{cpa} of them within 2 $\mathrm{m}$ are comparable, and the gap between them is less than 2\%. Also, with the generated \acs{rm}, the \acs{cpa} within 6 $\mathrm{m}$ and 10 $\mathrm{m}$ are over 70\% and 90\% respectively. This level of positioning accuracy is adequate for many applications requiring room level positioning accuracy (e.g., telling apart different shops in a big mall). The benefits of \acs{rm} generation method are in two aspects: i) it smooths the changes of \acs{rss} values caused by the temporal and spatial changes of the \acs{roi} (\figpref\ref{subfig:genrm}); and ii) it can generate the \acs{rm} for the region where is not covered by the site survey, it can thus reduce the labor of collecting the \acs{rm}.	
	\begin{table}[!htb]
		\caption{Results of \acs{svbi} based \acs{rm} generation}
		\label{tab:resultsofsvbi}
		\centering
		\begin{tabular}{c|c|c|c|c}
			\hline
			Results from&\multicolumn{2}{c}{\shortstack{Positioning \\ error ($\mathrm{m}$)}} \vline &\multicolumn{2}{c}{\shortstack{\acs{rss} estimation \\ error (dB)\footnote{The definition of \acs{rss} estimation error is as follows. Supposed the generated \acs{rss} $\hat{\mathbf{x}}$, the error comparing to its ground truth $ \mathbf{x} $ is: $ \sqrt{|\|\mathbf{x} - \hat{\mathbf{x}}\|_2^2/N_{\mathrm{AP}}} $}}}\\			
			\hline
			&mean &\acs{rmse}&mean &\acs{rmse}\\
			\hline
			\cite{Talvitie2015}\footnote{Because the proposed \acs{rm}generation algorithm in \cite{Talvitie2015} cannot achieve \acs{fbp}, they use \acs{knn} ($k=3$) to evaluate the performance.}& Appr. 5.5&--&10.2&--\\
			\cite{Majeed2016}&4.7&--&--&13\\
			\cite{Feng2012}\footnote{In \cite{Feng2012}, they employ affinity propagation based clustering and \acs{ap} selection methods before applying compressive sensing to \acs{fbp} and \acs{rm} generation. The result shown in \tabpref\ref{tab:resultsofsvbi} is the case that no clustering and \acs{ap} selection.}& Appr. 2.2&--&--&--\\
			This paper&\textbf{1.82} &\textbf{2.70}&\textbf{10.03}&\textbf{10.16}\\
			\hline
		\end{tabular}
	\end{table}

Comparing our results to the ones reported in \cite{Talvitie2015,Feng2012} and \cite{Majeed2016}, the proposed joint \acs{fbp} and \acs{rm} generation seems to provide better results in terms of both positioning and \acs{rss} estimation error (\tabpref\ref{tab:resultsofsvbi}). Regarding the positioning error, the mean error of the proposed approach is three times lower than that of distance based inter/extrapolation method in \cite{Talvitie2015} and more than two times lower than that of manifold alignment based method in \cite{Majeed2016}. The mean positioning error is 20\% lower than that of the compressive sensing based approach in \cite{Feng2012}. For the \acs{rss} estimation error, \acs{svbi} based and distance based inter/extrapolation methods obtain comparable performance regarding the mean error. Comparing to the manifold alignment based method, the proposed method herein achieves approximately 25\% reduction of \acs{rmse}. However, The comparison summarized in \tabpref\ref{tab:resultsofsvbi} is just an indication, since each of the results is based on different data and a different indoor environment.

\section{Conclusion}
Stochastic variational Bayesian inference (\acs{svbi}) is employed herein to accomplish joint \acf{fbp} and \acf{rm} generation. The performance in terms of both positioning and \acs{rss} estimation error of the proposed method is better than that of previous work. The proposed probabilistic model can be implemented based on \acfp{dnn} and trained jointly for both \acs{fbp} and \acs{rm} generation. Compared to the \acs{fbp} approaches based on the \acf{snn}, \acs{dnn} and \acf{knn}, the proposed \acs{svbi} based position estimation outperforms them. The reduction of \acf{rmse} of the localization is up to 40\% comparing to that of \acs{snn} based \acs{fbp}. And the \acl{cpa} within 4 $ \mathrm{m} $ of the proposed \acs{fbp} method is up to 92\% \acs{wrt} the positioning error, which is approximately 8\% higher than of \acs{knn} within 4 $ \mathrm{m} $. Regarding the performance of \acs{svbi} based \acs{rm} generation, it is comparable to that of the manually collect \acs{rm} and adequate for the applications, which require room level positioning accuracy.



\section*{Acknowledgment}
The China Scholarship Council (CSC) supports the first author during this doctoral studies and the second author as a visiting student at ETH Z\"urich.



\bibliographystyle{IEEEtran}
\bibliography{./reference/IPIN2017CZ}

\begin{thebibliography}{10}
\providecommand{\url}[1]{#1}
\csname url@samestyle\endcsname
\providecommand{\newblock}{\relax}
\providecommand{\bibinfo}[2]{#2}
\providecommand{\BIBentrySTDinterwordspacing}{\spaceskip=0pt\relax}
\providecommand{\BIBentryALTinterwordstretchfactor}{4}
\providecommand{\BIBentryALTinterwordspacing}{\spaceskip=\fontdimen2\font plus
\BIBentryALTinterwordstretchfactor\fontdimen3\font minus
  \fontdimen4\font\relax}
\providecommand{\BIBforeignlanguage}[2]{{%
\expandafter\ifx\csname l@#1\endcsname\relax
\typeout{** WARNING: IEEEtran.bst: No hyphenation pattern has been}%
\typeout{** loaded for the language `#1'. Using the pattern for}%
\typeout{** the default language instead.}%
\else
\language=\csname l@#1\endcsname
\fi
#2}}
\providecommand{\BIBdecl}{\relax}
\BIBdecl

\bibitem{Adler2015}
S.~Adler, S.~Schmitt, K.~Wolter, and M.~Kyas, ``{A survey of experimental
  evaluation in indoor localization research},'' \emph{Indoor Positioning and
  Indoor Navigation (IPIN), 2015 International Conference on}, no. October, pp.
  1--10, 2015.

\bibitem{1432143}
C.~Lee, Y.~Chang, G.~Park, J.~Ryu, S.-G. Jeong, S.~Park, J.~W. Park, H.~C. Lee,
  K.~shik Hong, and M.~H. Lee, ``Indoor positioning system based on incident
  angles of infrared emitters,'' in \emph{30th Annual Conference of IEEE
  Industrial Electronics Society, 2004. IECON 2004}, vol.~3, Nov 2004, pp.
  2218--2222 Vol. 3.

\bibitem{hazas2006broadband}
M.~Hazas and A.~Hopper, ``{Broadband ultrasonic location systems for improved
  indoor positioning},'' \emph{Mobile Computing, IEEE Transactions on}, vol.~5,
  no.~5, pp. 536--547, 2006.

\bibitem{Youssef2008}
M.~Youssef and A.~Agrawala, ``{The Horus location determination system},''
  \emph{Wireless Networks}, vol.~14, no.~3, pp. 357--374, 2008.

\bibitem{liu2007survey}
H.~Liu, H.~Darabi, P.~Banerjee, and J.~Liu, ``{Survey of wireless indoor
  positioning techniques and systems},'' \emph{Systems, Man, and Cybernetics,
  Part C: Applications and Reviews, IEEE Transactions on}, vol.~37, no.~6, pp.
  1067--1080, 2007.

\bibitem{1528431}
E.~Foxlin, ``{Pedestrian tracking with shoe-mounted inertial sensors},''
  \emph{IEEE Computer Graphics and Applications}, vol.~25, no.~6, pp. 38--46,
  nov 2005.

\bibitem{4167810}
H.~Wang, H.~Lenz, A.~Szabo, J.~Bamberger, and U.~D. Hanebeck, ``{WLAN-Based
  Pedestrian Tracking Using Particle Filters and Low-Cost MEMS Sensors},'' in
  \emph{2007 4th Workshop on Positioning, Navigation and Communication}, vol.
  2007, 2007, pp. 1--7.

\bibitem{Padmanabhan2000}
\BIBentryALTinterwordspacing
P.~B. Padmanabhan, V.~N., and V.~N., ``{RADAR: An in-building RF based user
  location and tracking system},'' \emph{Proceedings IEEE INFOCOM 2000.
  Conference on Computer Communications. Nineteenth Annual Joint Conference of
  the IEEE Computer and Communications Societies (Cat. No.00CH37064)}, vol.~2,
  no.~c, pp. 775--784, 2000. [Online]. Available:
  \url{http://research.microsoft.com/en-us/groups/sn-res/infocom2000.pdf}
\BIBentrySTDinterwordspacing

\bibitem{4907834}
V.~Honkavirta, T.~Perala, S.~Ali-Loytty, and R.~Piche, ``A comparative survey
  of wlan location fingerprinting methods,'' in \emph{2009 6th Workshop on
  Positioning, Navigation and Communication}, March 2009, pp. 243--251.

\bibitem{Ndrmyr2014}
S.~Niedermayr, A.~Wieser, and H.~Neuner, ``{E}xpressing location uncertainty in
  combined feature-based and geometric positioning,'' in \emph{Proceedings
  European Navigation Conference 2014}.\hskip 1em plus 0.5em minus 0.4em\relax
  s.l.: EUGIN, 2014, pp. 154--.

\bibitem{He2016}
S.~He and S.~H.~G. Chan, ``{Wi-Fi fingerprint-based indoor positioning: Recent
  advances and comparisons},'' \emph{IEEE Communications Surveys and
  Tutorials}, vol.~18, no.~1, pp. 466--490, 2016.

\bibitem{Zh2017}
C.~Zhou and A.~Wieser, ``{Application of backpropagation neural networks to
  both stages of fingerprinting based WIPS},'' in \emph{2016 Fourth
  International Conference on Ubiquitous Positioning, Indoor Navigation and
  Location Based Services (UPINLBS)}.\hskip 1em plus 0.5em minus 0.4em\relax
  Picataway, NJ: IEEE, 2017, pp. 207--217.

\bibitem{Xu2016}
\BIBentryALTinterwordspacing
J.~Xu, H.~Dai, and W.-h. Ying, ``{Multi-layer neural network for received
  signal strength-based indoor localisation},'' \emph{IET Communications},
  vol.~10, no.~6, pp. 717--723, 2016. [Online]. Available:
  \url{http://digital-library.theiet.org/content/journals/10.1049/iet-com.2015.0469}
\BIBentrySTDinterwordspacing

\bibitem{Zhang2016}
\BIBentryALTinterwordspacing
W.~Zhang, K.~Liu, W.~Zhang, Y.~Zhang, and J.~Gu, ``{Deep Neural Networks for
  wireless localization in indoor and outdoor environments},''
  \emph{Neurocomputing}, vol. 194, pp. 279--287, 2016. [Online]. Available:
  \url{http://dx.doi.org/10.1016/j.neucom.2016.02.055}
\BIBentrySTDinterwordspacing

\bibitem{Talvitie2015}
J.~Talvitie, M.~Renfors, and E.~S. Lohan, ``{Distance-based interpolation and
  extrapolation methods for RSS-based localization with indoor wireless
  signals},'' \emph{IEEE Transactions on Vehicular Technology}, vol.~64, no.~4,
  pp. 1340--1353, 2015.

\bibitem{Atia2013}
M.~M. Atia, A.~Noureldin, and M.~J. Korenberg, ``{Dynamic online-calibrated
  radio maps for indoor positioning in wireless local area networks},''
  \emph{IEEE Transactions on Mobile Computing}, vol.~12, no.~9, pp. 1774--1787,
  2013.

\bibitem{Feng2012}
C.~Feng, W.~S.~A. Au, S.~Valaee, and Z.~Tan, ``{Received-signal-strength-based
  indoor positioning using compressive sensing},'' \emph{IEEE Transactions on
  Mobile Computing}, vol.~11, no.~12, pp. 1983--1993, 2012.

\bibitem{Majeed2016}
K.~Majeed, S.~Sorour, T.~Y. Al-Naffouri, and S.~Valaee, ``{Indoor localization
  and radio map estimation using unsupervised manifold alignment with geometry
  perturbation},'' \emph{IEEE Transactions on Mobile Computing}, vol.~15,
  no.~11, pp. 2794--2808, 2016.

\bibitem{Demuth2014}
H.~B. Demuth, M.~H. Beale, O.~De~Jess, and M.~T. Hagan, \emph{Neural network
  design}.\hskip 1em plus 0.5em minus 0.4em\relax Martin Hagan, 2014.

\bibitem{Vincent2010}
P.~Vincent, H.~Larochelle, I.~Lajoie, Y.~Bengio, and P.-A. Manzagol, ``Stacked
  denoising autoencoders: Learning useful representations in a deep network
  with a local denoising criterion,'' \emph{Journal of Machine Learning
  Research}, vol.~11, no. Dec, pp. 3371--3408, 2010.

\bibitem{LeCun2012}
Y.~A. LeCun, L.~Bottou, G.~B. Orr, and K.~R. M\"uller, ``{Efficient
  backprop},'' \emph{Lecture Notes in Computer Science (including subseries
  Lecture Notes in Artificial Intelligence and Lecture Notes in
  Bioinformatics)}, vol. 7700 LECTU, pp. 9--48, 2012.

\bibitem{lee2008sparse}
H.~Lee, C.~Ekanadham, and A.~Y. Ng, ``Sparse deep belief net model for visual
  area v2,'' in \emph{Advances in neural information processing systems}, 2008,
  pp. 873--880.

\bibitem{hinton2006reducing}
G.~E. Hinton and R.~R. Salakhutdinov, ``Reducing the dimensionality of data
  with neural networks,'' \emph{science}, vol. 313, no. 5786, pp. 504--507,
  2006.

\bibitem{pulkkinen2011semi}
T.~Pulkkinen, T.~Roos, and P.~Myllym{\"a}ki, ``Semi-supervised learning for
  wlan positioning,'' in \emph{International Conference on Artificial Neural
  Networks}.\hskip 1em plus 0.5em minus 0.4em\relax Springer, 2011, pp.
  355--362.

\bibitem{Kingma2013}
\BIBentryALTinterwordspacing
D.~P. Kingma and M.~Welling, ``{Auto-Encoding Variational Bayes},'' \emph{arXiv
  preprint arXiv:1312.6114}, no.~Ml, pp. 1--14, 2013. [Online]. Available:
  \url{http://arxiv.org/abs/1312.6114}
\BIBentrySTDinterwordspacing

\bibitem{Kingma2015}
D.~P. Kingma and J.~L. Ba, ``{Adam: a Method for Stochastic Optimization},''
  \emph{International Conference on Learning Representations 2015}, pp. 1--15,
  2015.

\bibitem{Schulman2015}
\BIBentryALTinterwordspacing
J.~Schulman, N.~Heess, T.~Weber, and P.~Abbeel, ``{Gradient Estimation Using
  Stochastic Computation Graphs},'' \emph{Nips}, pp. 1--13, 2015. [Online].
  Available: \url{http://arxiv.org/abs/1506.05254}
\BIBentrySTDinterwordspacing

\bibitem{Paisley2012}
\BIBentryALTinterwordspacing
J.~Paisley, D.~Blei, and M.~Jordan, ``{Variational Bayesian Inference with
  Stochastic Search},'' \emph{Icml}, no. 2000, pp. 1367--1374, 2012. [Online].
  Available: \url{http://icml.cc/2012/papers/687.pdf}
\BIBentrySTDinterwordspacing

\bibitem{Rezende2014}
D.~J. Rezende, S.~Mohamed, and D.~Wierstra, ``{Stochastic backpropagation and
  approximate inference in deep generative models},'' \emph{JMLR: W\&CP},
  vol.~32, pp. 1278--1286, 2014.

\bibitem{Devroye1986}
\BIBentryALTinterwordspacing
L.~Devroye, ``{Sample-based Non-uniform Random Variate Generation},'' in
  \emph{Proceedings of the 18th Conference on Winter Simulation}, ser. WSC
  '86.\hskip 1em plus 0.5em minus 0.4em\relax New York, NY, USA: ACM, 1986, pp.
  260--265. [Online]. Available: \url{http://doi.acm.org/10.1145/318242.318443}
\BIBentrySTDinterwordspacing

\bibitem{IMM2012-03274}
\BIBentryALTinterwordspacing
K.~B. Petersen and M.~S. Pedersen, ``The matrix cookbook,'' nov 2012, version
  20121115. [Online]. Available: \url{http://www2.imm.dtu.dk/pubdb/p.php?3274}
\BIBentrySTDinterwordspacing

\bibitem{Bishop:2006:PRM:1162264}
C.~M. Bishop, \emph{Pattern Recognition and Machine Learning (Information
  Science and Statistics)}.\hskip 1em plus 0.5em minus 0.4em\relax Secaucus,
  NJ, USA: Springer-Verlag New York, Inc., 2006.

\bibitem{Sohn2015}
K.~Sohn, H.~Lee, and X.~Yan, ``{Learning Structured Output Representation using
  Deep Conditional Generative Models},'' \emph{Advances in Neural Information
  Processing Systems}, pp. 3483--3491, 2015.

\bibitem{Kingma2014}
\BIBentryALTinterwordspacing
D.~Kingma, D.~Rezende, and M.~Welling, ``{Semi-supervised Learning with Deep
  Generative Models},'' \emph{arXiv preprint}, pp. 1--9, 2014. [Online].
  Available: \url{http://arxiv.org/abs/1406.5298}
\BIBentrySTDinterwordspacing

\bibitem{tensorflow2015-whitepaper}
\BIBentryALTinterwordspacing
M.~Abadi, A.~Agarwal, P.~Barham, E.~Brevdo, Z.~Chen, C.~Citro, G.~S. Corrado,
  A.~Davis, J.~Dean, M.~Devin, S.~Ghemawat, I.~Goodfellow, A.~Harp, G.~Irving,
  M.~Isard, Y.~Jia, R.~Jozefowicz, L.~Kaiser, M.~Kudlur, J.~Levenberg,
  D.~Man\'{e}, R.~Monga, S.~Moore, D.~Murray, C.~Olah, M.~Schuster, J.~Shlens,
  B.~Steiner, I.~Sutskever, K.~Talwar, P.~Tucker, V.~Vanhoucke, V.~Vasudevan,
  F.~Vi\'{e}gas, O.~Vinyals, P.~Warden, M.~Wattenberg, M.~Wicke, Y.~Yu, and
  X.~Zheng, ``{TensorFlow}: Large-scale machine learning on heterogeneous
  systems,'' 2015, software available from tensorflow.org. [Online]. Available:
  \url{http://tensorflow.org/}
\BIBentrySTDinterwordspacing

\bibitem{Gu2017}
Y.~Gu, C.~Zhou, A.~Wieser, and Z.~Zhou, ``{Pedestrian Positioning Using WiFi
  Fingerprints and A Foot-mounted Inertial Sensor},'' \emph{arXiv preprint
  arXiv:1704.03346}, no.~1, pp. 1--9, 2017.

\bibitem{Glorot2010}
X.~Glorot and Y.~Bengio, ``{Understanding the difficulty of training deep
  feedforward neural networks},'' \emph{Proceedings of the 13th International
  Conference on Artificial Intelligence and Statistics (AISTATS)}, vol.~9, pp.
  249--256, 2010.

\bibitem{Peters2001}
C.~A. Peters, ``Statistics for analysis of experimental data,''
  \emph{Environmental Engineering Processes Laboratory Manual}, pp. 1--25,
  2001.

\bibitem{scikit-learn}
F.~Pedregosa, G.~Varoquaux, A.~Gramfort, V.~Michel, B.~Thirion, O.~Grisel,
  M.~Blondel, P.~Prettenhofer, R.~Weiss, V.~Dubourg, J.~Vanderplas, A.~Passos,
  D.~Cournapeau, M.~Brucher, M.~Perrot, and E.~Duchesnay, ``Scikit-learn:
  Machine learning in {P}ython,'' \emph{Journal of Machine Learning Research},
  vol.~12, pp. 2825--2830, 2011.

\end{thebibliography}
%
%
%

\end{document}